\documentclass{article}

\PassOptionsToPackage{numbers, compress}{natbib}

\usepackage[preprint]{neurips_2019}

\usepackage[utf8]{inputenc} 
\usepackage[T1]{fontenc}    
\usepackage{hyperref}       
\usepackage{url}            
\usepackage{booktabs}       
\usepackage{amsfonts}       
\usepackage{nicefrac}       
\usepackage{microtype}      
\usepackage{relsize}
\usepackage{wrapfig}
\usepackage{url}

\usepackage{amsmath}
\usepackage[ruled]{algorithm2e}
\usepackage{graphicx}
\usepackage{gensymb}
\usepackage{pifont}
\usepackage{dsfont}
\usepackage{floatrow}
\usepackage{mathtools}
\newfloatcommand{capbtabbox}{table}[][\FBwidth]
\newcommand*{\bs}{\boldsymbol}
\DeclareMathOperator*{\argmax}{arg\,max}

\newcommand*\diff{\mathop{}\!\mathrm{d}}
\newcommand{\expect}{\mathbb{E}}
\mathtoolsset{showonlyrefs}
\newtheorem{theorem}{Theorem}

\title{Ensemble Model Patching: A Parameter-Efficient Variational Bayesian Neural Network}

\author{
{\bf Oscar Chang$\dag$,~~ Yuling Yao$\ddag$, ~~ David Williams-King$\dag$,~~  Hod Lipson$\dag$}\\
$\dag$Department of Computer Science, ~~ $\ddag$ Department of Statistics\\
Columbia University\\
\texttt{\small\{oscar.chang, yy2619, hod.lipson\}@columbia.edu, 
dwk@cs.columbia.edu}
\vspace{-0.5em}
}

\begin{document}

\maketitle
\begin{abstract}
Two main obstacles preventing the widespread adoption of variational Bayesian neural networks are the high parameter overhead that makes them infeasible on large networks, and the difficulty of implementation, which can be thought of as ``programming overhead." MC dropout \citep{gal2016dropout} is popular because it sidesteps these obstacles. Nevertheless, dropout is often harmful to model performance when used in networks with batch normalization layers \citep{li2018understanding}, which are an indispensable part of modern neural networks. We construct a general variational family for ensemble-based Bayesian neural networks that encompasses dropout as a special case. We further present two specific members of this family that work well with batch normalization layers, while retaining the benefits of low parameter and programming overhead, comparable to non-Bayesian training. Our proposed methods improve predictive accuracy and achieve almost perfect calibration on a ResNet-18 trained with ImageNet.
\end{abstract}

\section{Introduction}
\label{section:intro}


As deep learning becomes ubiquitous in safety-critical applications like autonomous driving and medical imaging, it is important for neural network practitioners to quantify the degree of belief they have in their predictions \citep{moosavi2016deepfool, amodei2016concrete,zhang2016understanding}. Unlike conventional deep neural networks which are poorly calibrated \citep{quia2010sparse, goodfellow2014explaining,  guo2017calibration, lakshminarayanan2017simple}, Bayesian neural networks learn a probability distribution over parameters.
This design enables uncertainty estimation, allows for better-calibrated probability prediction, and reduces overfitting. 

However, exact posterior inference for deep Bayesian neural networks is intractable in general, so approximate methods like variational inference are often used \citep{gal2016dropout,graves2011practical,hernandez2015probabilistic,blundell2015weight,louizos2016structured,louizos2017multiplicative,kingma2015variational}. Unfortunately, most of the proposed variational methods still require significant (i.e.\ $\geq$100\%) parameter overhead and do not scale well to modern neural networks with millions of parameters. For example, a mean-field Gaussian \citep{graves2011practical,blundell2015weight} 
\emph{doubles} the parameter use (by learning both means and variances). These techniques also incur significant \emph{programming overhead} since the programmer must perform extensive changes to their neural network architecture to make it Bayesian. For example, \citep{hernandez2015probabilistic,blundell2015weight} require a modification to the backpropagation algorithm, while \citep{louizos2016structured,louizos2017multiplicative,krueger2017bayesian} involve complicated weight-sampling techniques.

\begin{figure}
\CenterFloatBoxes
\footnotesize
\begin{floatrow}
\ffigbox[.44\textwidth]{
    \includegraphics[width=0.5\textwidth]{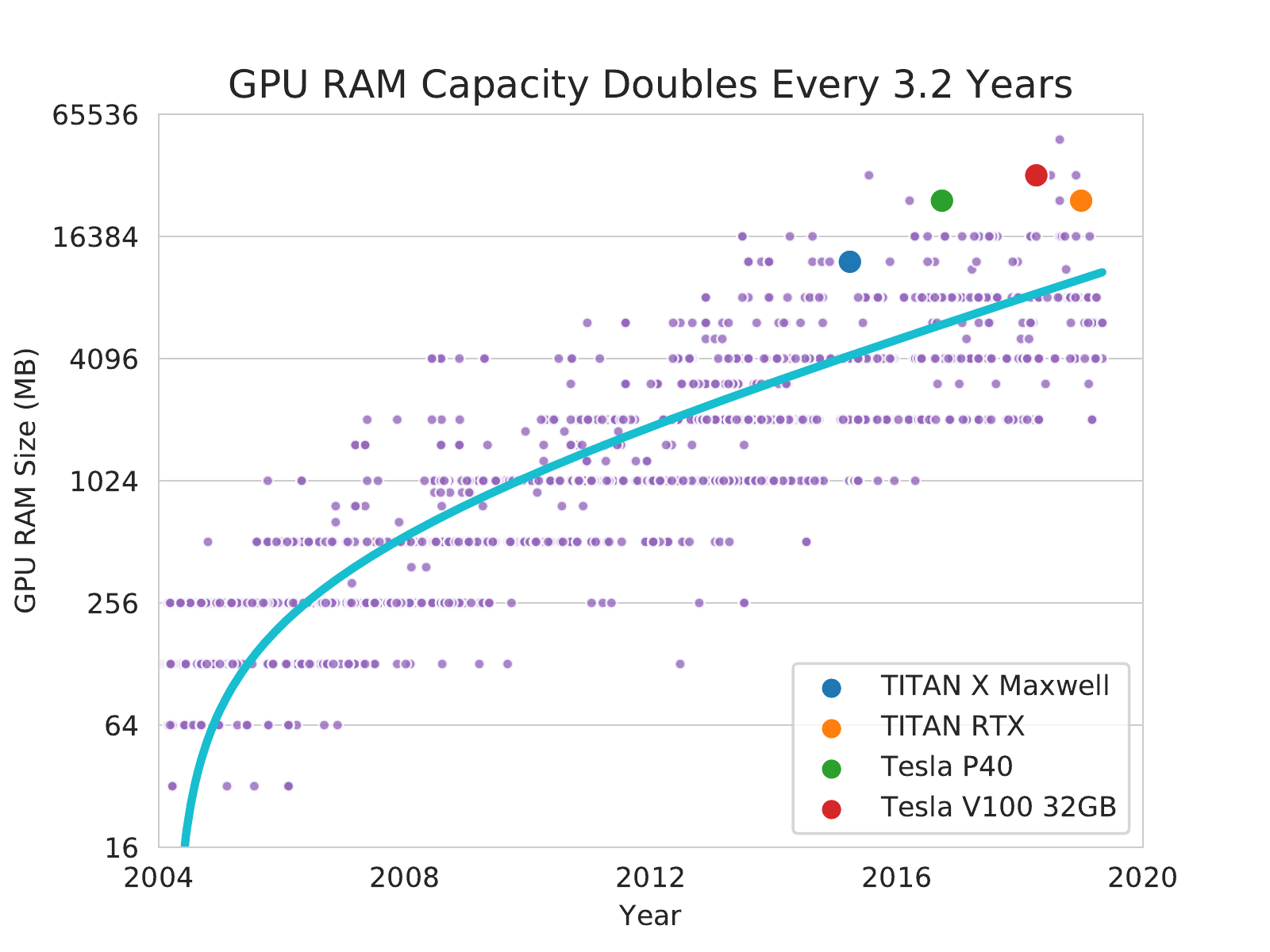}
    \centering
       \vspace{-1em}
    }{%
    \vspace{-1em}
    \caption{\footnotesize GPU RAM capacity grows exponentially, doubling every $3.2$ years for the last $15$ years.
    We scraped the data for $1599$ GPUs and fitted a log-linear model; details in Appendix~\ref{section:gpu_analysis}.
    }
    \label{fig:gpu_ram}
}
\killfloatstyle
\capbtabbox[.56\textwidth]{%
    \begin{tabular}{p{2.38cm} p{1.6cm} p{1.6cm}}
     \toprule
     Variational\newline Method & ResNet-18 Parameters  \footnotesize{[\emph{Overhead}]} & PyramidNet Parameters  \footnotesize{[\emph{Overhead}]}\\
     \midrule
     Mean-Field\newline Gaussian \citep{graves2011practical,blundell2015weight} & $23.4 M$ [\emph{100\%}] & $57.0M$ [\emph{100\%}]\\
     PBP \citep{hernandez2015probabilistic} & $23.4M$ [\emph{100\%}] & $57.0M$ [\emph{100\%}]\\
     MNFG \citep{louizos2017multiplicative} & $3,510M$ [\emph{29900\%}] & $8,550M$ [\emph{29900\%}]\\
     Dropout \citep{gal2016dropout,kingma2015variational} & $11.7M$ [\emph{0.00\%}] & $28.5M$ [\emph{0.00\%}]\\
     {Ensemble Model\newline Patching (ours)} & $13.8M$ [\emph{17.9\%}] & $28.8M$ [\emph{1.10\%}]\\
     \bottomrule
    \end{tabular}
}{%
  \caption{\footnotesize Variational methods in the literature incur significant parameter overhead. We computed these parameter counts 
  based on recommended hyper-parameter settings. 
  }%
  \label{table:vbnn_params}\vspace{-0.8em}
}
\end{floatrow}
\vspace{-0.5cm}
\end{figure}
\normalsize 
High parameter overhead is an important concern,
as deep learning models already utilize available hardware resources to the fullest, often maxing out GPU memory usage \citep{hanlon2017memory}.
In 2015, the highest end consumer-grade GPU 
had 12GB of memory;
by 2018, this number had doubled to 24GB. 
This means that in the period 2015-2018, a programmer could only have Bayesianized state-of-the-art neural networks from prior to 2015. Any newer designs, once their parameter use had been doubled, would not fit into her GPU's memory.
With GPU memory capacity doubling approximately every $3.2$ years (Figure \ref{fig:gpu_ram})---an eternity in the rapidly progressing field of deep learning---the programmer's Bayesianized networks will be up to three years behind the state of the art.

We survey in Table \ref{table:vbnn_params} the parameter use of several existing Variational Bayesian Neural Network (VBNN) methods for ResNet-18 \citep{he2016deep} and PyramidNet \citep{han2017deep},
among which dropout is the only method that does not add a significant ($\geq$100\%) parameter overhead. 
Nonetheless, in the presence of batch normalization, a mainstay of modern deep networks, dropout layers have been found to be either redundant \citep{ioffe2015batch}, or downright harmful to model performance due to shifts in variance \citep{li2018understanding}.

In this paper, our contributions are two-fold.
First, we construct a general variational family for ensemble-based Bayesian neural networks.
This unified family extends and bridges the Bayesian interpretation for both implicit and explicit ensembles, where methods like dropout, DropConnect \citep{wan2013regularization} and explicit ensembling \citep{lakshminarayanan2017simple, pearce2018bayesian} can be viewed as special cases. We show 
the low parameter overhead and large ensemble sizes for several implicit ensemble methods, thus suggesting their suitability for use in large neural networks.

Second, we present two novel variational distributions (Ensemble Model Patching) for implicit ensembles. They are better alternatives to dropout especially in batch-normalized networks. Our proposed methods outperform MC dropout and non-Bayesian networks on test accuracy, probability calibration, and robustness against common image corruptions. The parameter overhead and computational cost are nearly the same as in non-Bayesian training, making Bayesian networks affordable on large scale datasets and modern deep residual networks with hundreds of layers. To our knowledge, we are the first to scale a Bayesian neural network to the ImageNet dataset, achieving almost perfect calibration.
Our methods are both scalable and easy to implement, serving as one-to-one replacements for particular layers in a neural network without further changes to its architecture.

The remainder of the paper is organized as follows. We propose a general variational distribution for ensemble-based VBNNs in Section \ref{section:ensemble_vbnn}, before introducing Ensemble Model Patching and its two variants in Section \ref{section:emp}.
We discuss related work in Section \ref{section:related_work}, validate our proposed methods experimentally in Section \ref{section:experiments}, and finally conclude our findings in Section \ref{section:conclusion}. For an introduction to technical background, we refer readers to Appendix \ref{section:background}. We provide derivation details and theoretical proofs in Appendix \ref{section:klmog}. More implementation details can be found in Appendices \ref{section:parallelization} and \ref{section:moreexp}.

\section{Ensemble-Based Variational Bayesian Neural Networks}
\label{section:ensemble_vbnn}
\paragraph{Implicit vs Explicit Ensembles}
A classical ensemble, which we call an explicit ensemble in this paper, involves fitting several different models and combining the output using a method like averaging or majority vote. An explicit ensemble with $K$ components thus has to maintain $K$ models, which is inefficient. Implicit ensembles, by contrast, are more efficient because the different components arise through varying parts of each model rather than the whole model. For example, the Bernoulli-Gaussian model \citep{murphy2012machine}, which can be considered a conceptual predecessor to dropout, maintains $D$ products of a Bernoulli and a Gaussian random variable. Thus, each realization of the Bernoulli results in a different model, leading implicitly to $2^D$ mixture components.

Can we unify these two approaches with a continuous expansion \citep{gelman2013philosophy}? Yes, assuming that all the ensemble components have the same network architecture, we can indeed model both explicit and implicit neural network ensembles as variational Bayes.
This general variational family, which includes MC dropout as a specific case, lends a Bayesian interpretation to both implicit ensembles like DropConnect \citep{wan2013regularization} and also explicit ensembles like \citet{lakshminarayanan2017simple}'s Deep Ensembles.

\paragraph{A General Variational Family}
In Equation \eqref{eq_variationalfamily}, we construct a general  variational family with a factorial distribution over mixtures of Gaussians for both the network's weights $\textbf{W}\!=\{\mathbf{W}_{i}\}_{i=1}^L$ and biases $\textbf{b}\!=\mathbf\{\mathbf{b}_j\}_{j=1}^{L}$, where $i, j$ are layer indices, $L$ is the number of layers, $K_i$ is the number of mixture components in the $i$-th layer, and $H_{i}$ is the output dimension of the $i$-th layer. $\mathbf{Z}_i \in \{1,\dots, K_i\}^{H_{i-1} \times H_i}$ and $\mathbf{z}_j \in \{1,\dots, K_j\}^{H_{j}}$ are categorical variables that indicate the assignment among  mixture components and are generated from a Bernoulli or multinoulli distribution with probability $p_i$ or $\mathbf{p}\!=\!\{{p}_{ik}\!\}$. $\odot$ denotes the Hadamard product and $\Theta=(\mathbf W, \mathbf b)$ the set of parameters.
\begin{equation}\label{eq_variationalfamily}
\begin{split}
\mathbf{W}_{i}&= \sum_{k=1}^{K_i} \mathds{1}[\mathbf{Z}_i = k] \odot (\mathbf{M}_{ik}+{\sigma} \mathbf{\epsilon}_{ik}), \qquad
\mathbf{\epsilon}_{ik}  \sim \mathcal{N}(\mathbf{0}, \mathbf{I}_{H_{i-1} \times H_i}), \quad
1 \leq i  \leq L, 
\\
\mathbf{b}_j&= \sum^{K_j}_{k=1} \mathds{1}[\mathbf{z}_j = k] \odot (\mathbf{m}_{jk}+{\sigma} \mathbf{\epsilon}_{jk}),  \qquad
\mathbf{\epsilon}_{jk} \sim \mathcal{N}(\mathbf{0}, \mathbf{I}_{H_j}), \qquad
  1 \leq j  \leq L.\\
\end{split}
\end{equation}
Throughout this paper, the categorical mixing probabilities $\mathbf{p}$ are fixed to be uniform over all $K$ ensemble components for simplicity. If $\mathbf{Z}$ is marginalized out, the variational distribution is fully parametrized by $(\mathbf{M}=\{\mathbf{M}_{ik}\}_{k=1,  i=1}^{K_i,L}$, $\mathbf{m}=\{\mathbf{m}_{jk}\}^{K_j, L}_{k=1, j=1})$. They are centroids of the mixing components of \textbf{W}  and \textbf{b}, and are parameters in each component of the neural network ensemble.


We show examples of specific members of this family in Table \ref{table:eb_vbnns}.
\vspace{-0.5em}
\begin{table}[h]
\caption{Special Cases of Variational Distributions for Different Ensemble-Based VBNNs}
\vspace{-0.5em}
\centering
\footnotesize
\begin{tabular}{p{2.2cm} p{3.2cm} p{3.2cm} p{4cm} }
 \toprule
  & \textbf{Dropout} & \textbf{DropConnect} & \textbf{Explicit Ensemble}\\
 \midrule
Number of \newline Components & $K_i = 2, K_j = 1$ & $K_i = 2, K_j = 1$ &   $K_i = K_j = K \geq 2$\\
\vspace{0.01em} Mixing\newline Assignment
   & $\mathbf{z}_{i h_1} \sim \text{Bernoulli}(p_i)$\newline
     $\mathbf{Z}_{i, h_1 h_2} = z_{i h_1}$
   & $\mathbf{Z}_{i, h_1 h_2} \sim \text{Bernoulli}(p_i)$
   & $z \sim \text{Categorical}(\mathbf{p})$\newline
     $\mathbf{Z}_{i, h_1 h_2} = z, $
     $\mathbf{z}_{j, h_3} = z$\\
 \addlinespace[0.1cm]
 \vspace{0.04em} Constants
   & $\mathbf{M}_{i2} = \mathbf{0}_{H_{i-1} \times H_i}$\newline
     $\mathbf{z}_j = \mathbf{1}_{H_j}$
   & $\mathbf{M}_{i2} = \mathbf{0}_{H_{i-1} \times H_i}$\newline
     $\mathbf{z}_j = \mathbf{1}_{H_j}$
   & N/A\\
 \addlinespace[0.1cm]
 Variational Parameters 
   & $\{p_i\}_{i=1}^{L}$ \newline $\{\mathbf{M}_{i1}\}_{i=1}^{L}$,  $\{\mathbf{m}_{j1}\}^{L}_{j=1}$
   & $\{p_i\}_{i=1}^{L}$ \newline $\{\mathbf{M}_{i1}\}_{i=1}^{L}$, $\{\mathbf{m}_{j1}\}^{L}_{j=1}$
   & $\mathbf{p}$\newline
   $\{\mathbf{M}_{ik}\}_{k=1,i=1}^{K, L}$, $\{\mathbf{m}_{jk}\}^{K, L}_{k=1,j=1}$\\
 \bottomrule
\end{tabular}
\label{table:eb_vbnns}
\vspace{-0.5em}
\end{table}
\normalsize 

\paragraph{Evaluating the Evidence Lower Bound (ELBO)}
In all cases, we set the prior on weights $p^{\mathrm{prior}}(\Theta)$ to be a zero-centered isotropic Gaussian with precision $\tau$ and $\rho$, and evaluate the Kullback–Leibler (KL) divergence in the ELBO as follows (for further details, see Appendix \ref{section:klmog}):
\begin{equation}
\label{eqn:kl_general}
\mathrm{KL}\left(q(\Theta | \mathbf{M,m} )||\ p^{\mathrm{prior}}(\Theta)\right) \approx \sum_{i=1}^{L} \sum_{k=1}^{K_i} \frac{\tau_i p_{ik}}{2} || \mathbf{M}_{ik} ||^2_2 + \sum_{j=1}^{L} \sum_{k=1}^{K_j} \frac{\rho_j p^*_{jk}}{2} || \mathbf{m}_{jk} ||^2_2 + \mathrm{Constant}.
\end{equation}
\begin{equation}
\text{where}\ p_{ik} = 
\begin{cases}
p_i & \text{for Dropout and DropConnect}\\
\mathbf{p}_{k} & \text{for Explicit Ensemble}
\end{cases}
\ \ p^*_{jk} = 
\begin{cases}
1 & \text{for Dropout and DropConnect}\\
\mathbf{p}_{k} & \text{for Explicit Ensemble}
\end{cases}
\end{equation}
The approximation \eqref{eqn:kl_general} is reasonable as long as the individual ensemble components do not significantly overlap, which will be satisfied when the dimension of $\Theta$ is high. A practical implementation is to simply enforce L2 regularization on all the learnable variational parameters.

\section{Ensemble Model Patching}
\label{section:emp}
Through Bernoulli mixing, dropout fixes a component centered at zero. The zeros are harmful to a batch-normalized network as shown in \citep{li2018understanding} and confirmed by our experiments. We thus seek a method where the weights in each layer mix over several ensemble components and all components are simultaneously optimized. This is prohibitively expensive if we apply it to the entire network, but since deep neural networks are over-parametrized, we can target the small fraction of weights (model patches) that have a disproportionate effect---a technique first described by \citet{mudrakarta2018k}.

\subsection{Partitioning the Variational Distribution}
A \emph{model patch} \citep{mudrakarta2018k} refers to a small subset of a neural network, which can be substituted for task-adapted weights in multi-task and transfer learning. We use $\Lambda_p$ to denote the set of indices for patched layers and $\Lambda_s$ for shared layers. This divides the set of all network parameters $\Theta$ into patched parameters $\Theta_{\mathrm{patch}}= (\{\mathbf{W}_{i}\}_{i \in \Lambda_p}, \{\mathbf{b}_{j}\}_{j \in \Lambda_p}  )$  and shared ones $\Theta_{\mathrm{shared}}= (\{\mathbf{W}_{i}\}_{i \in \Lambda_s}, \{\mathbf{b}_{j}\}_{j \in \Lambda_s})$.


To reduce the parameter overhead, we construct the variational distribution as a product of shared and patched parameters separately, $q(\Theta)= \!q(\Theta_{\mathrm{shared}})q(\Theta_{\mathrm{patch}})$, where $q(\Theta_{\mathrm{patch}})$ is an ensemble-based distribution from Equation \eqref{eq_variationalfamily}, and $q(\Theta_{\mathrm{shared}})$ is a mean-field Gaussian distribution
\begin{equation}\label{eq_shared}
q(\Theta_{\mathrm{shared}}|{\mathbf{M, m}} )
=\prod_{i \in \Lambda_s  } q(\textbf{W}_{i} ) q(\textbf{b}_{i})  
= \prod_{i \in \Lambda_s  } \mathcal{N}(\textbf{W}_{i}| \textbf{M}_i, \sigma^2 \textbf{I})  \mathcal{N}(\textbf{b}_{i}| \textbf{m}_i, \sigma^2 \textbf{I}).
\end{equation}
We further simplify the variational distribution by fixing $\sigma$ to be small, e.g.\ machine epsilon.

\subsection{Proposed Algorithms: EMP and ECMP} 
\begin{figure}
\CenterFloatBoxes
\footnotesize
\begin{floatrow}
\ffigbox{%
    \includegraphics[width=0.5\textwidth]{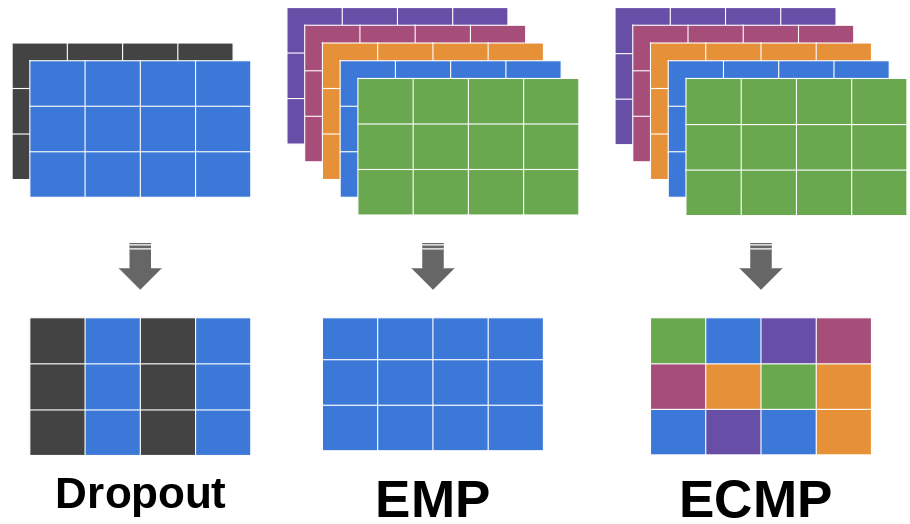}
    \centering
    }{%
    \caption{\footnotesize Assignment of weights using different schemes. Dropout draws each \emph{column} of the weight matrix either from the weight or zero component. EMP samples each patch layer jointly, thus the weight \emph{matrix} is from a single component. ECMP draws each weight in a layer from an ensemble, thus each \emph{element} in a weight matrix can belong to different components.}
    \label{fig:emp_v_ecmp}
}
\killfloatstyle
\ffigbox{%
\begin{algorithm}[H]
\SetAlgoLined
\caption{\footnotesize A forward pass with Ensemble Model Patching. We initialize \textbf{M} and \textbf{m} randomly before the start of training. Input is $x$, $S$ is the number of MC draws; $S=1$ for training in our experiments.}%
 \label{alg:emp_pseudocode}
 $\hat{y} := 0$;\\
 \For{\emph{[MC sample index]} $s\gets1$ \KwTo $S$}{
    $y_s := x$;\\
    \For{\emph{[Layer index]} $i\gets1$ \KwTo $L$ }{
        \uIf{$i \in \Lambda_{\mathrm{patch}}$}{
            Sample mixing assignment $\mathbf{Z}_i$\;
            Sample $\Theta_i$ and thus $\text{Layer}_i$ through EMP \eqref{eq_emp} or ECMP \eqref{eq_ecmp} and \eqref{eq_variationalfamily}\;
        }
         \uElseIf{$i \in \Lambda_{\mathrm{shared}}$}{
         Sample $\Theta_i$ and thus $\text{Layer}_i$ through \eqref{eq_shared}\;
        }
    $y_s := \text{Layer}_i (y_s)$\;
    }
    $\hat{y} := \hat{y} + (y_s - \hat{y})/(s+1)$\;
}
\Return $\hat{y}$
\end{algorithm}
}{}
\end{floatrow}
\vspace{-1em}
\end{figure}

We identify two variational distributions for $q(\Theta_{\mathrm{patch}})$ that avoid the hard zeros in dropout while retaining its low parameter and programming overhead. 
We fix the number of ensembles \emph{in each layer} to be $K$,  and we write the \textbf{Ensemble Model Patching (EMP)} distribution as follows. 
\begin{equation}\label{eq_emp}
\begin{split}
K_i &= K \geq 2, \ \  z_i \sim\ \text{Categorical}(\mathbf{p}_i),
\text{for}\ i \in \Lambda_p,\\
\mathbf{Z}_{i, h_1 h_2} &= z_i\ \text{for}\ h_1 \in [1, H_{i-1}], h_2 \in [1, H_{i}], i \in \Lambda_p,\\
\mathbf{z}_{j, h_3} &= z_j\ \text{for}\ h_3 \in [1, H_{j}], j \in \Lambda_p.\\
\end{split}
\end{equation}
In \eqref{eq_emp}, given $i$, the matrix $\mathbf{Z}_{i, h_1 h_2}$ remains the same for all elements $(h_1, h_2)$.  
Instead, we can sample each element in $\mathbf{Z}_{i, h_1 h_2}$, $\mathbf{z}_{j, h_3}$  independently. We call this variant \textbf{Ensemble Cross Model Patching (ECMP)}; see Figure \ref{fig:emp_v_ecmp} for a visual illustration of the distinction between the two methods. In ECMP,
\begin{equation}\label{eq_ecmp}
\begin{split}
K_i &= K \geq 2, \text{for}\ i \in \Lambda_p,\\
\mathbf{Z}_{i, h_1 h_2} &\sim \text{Categorical}(\mathbf{p}_i)\ \text{for}\ h_1 \in [1, H_{i-1}], h_2 \in [1, H_{i}], i \in \Lambda_p,\\
\mathbf{z}_{j, h_3} &\sim \text{Categorical}(\mathbf{p}_j)\ \text{for}\ h_3 \in [1, H_{j}], j \in \Lambda_p.\\
\end{split}
\end{equation}
ECMP masks the hidden weight matrix, analogous to DropConnect, but avoids its hard zeros. Combining \eqref{eq_emp} or \eqref{eq_ecmp} with \eqref{eq_variationalfamily}, we obtain the complete variational distribution for $q(\Theta_{\mathrm{patch}})$. In both cases, the variational parameters to optimize over are:
$$\left( \{\mathbf{M}_{ik}\}_{k=1,{i\in \Lambda_p}}^{K_i}, 
 \{\mathbf{m}_{jk}\}_{k=1,{j\in \Lambda_p}}^{K_j},
  \{\mathbf{M}_{i}\}_{{i\in \Lambda_s}}, 
 \{\mathbf{m}_{j}\}_{{j\in \Lambda_s}}\right).$$
We set the prior on all parameters to be a zero-centered isotropic Gaussian: 
$$p^{\mathrm{prior}}(\Theta)= \prod_{i=1}^L \mathcal{N}( \mathbf{W}_i| \mathbf{0}, \tau_i^{-1} \mathbf{I}) \prod_{j=1}^{L} \mathcal{N}( \mathbf{b}_j| \mathbf{0}, \rho_j^{-1} \mathbf{I}).$$ Applying Equation \eqref{eqn:kl_general} to the patch layers, we approximate the $\mathrm{KL}\left(q(\Theta \mid \mathbf{M,m})||\ p^{\mathrm{prior}}(\Theta)\right)$  (up to an additive constant) for both EMP and ECMP as
\begin{equation}
\label{eqn:kl_emp}
 \sum_{i\in \Lambda_p} \sum_{k=1}^{K_i} \frac{\tau_i p_{ik}}{2} || \mathbf{M}_{ik} ||^2_2 + \sum_{j \in \Lambda_p} \sum_{k=1}^{K_j} \frac{\rho_j p_{jk}}{2} || \mathbf{m}_{jk} ||^2_2 +  \sum_{i\in \Lambda_s}\! \frac{\tau_i}{2} || \mathbf{M}_{i} ||^2_2 + \sum_{j \in \Lambda_s}\!\!  \frac{\rho_j }{2} || \mathbf{m}_{j} ||^2_2. 
\end{equation}
The final loss function (negative ELBO) is then \eqref{eqn:kl_emp} minus the Monte Carlo (MC) estimation of the likelihood $1/S \sum_{s=1}^S  \log p(\mathbf{Y} | \mathbf{X}, \Theta_s)$, where $\Theta_s$ is the $s$-th MC draw of $\Theta$. In our experiments, we set $S=1$ for gradient evaluation, so it becomes the conventional squared error or cross entropy loss for outcome $y$. Details for the derivation can be found in Appendix \ref{section:klmog}. Note that the term \eqref{eqn:kl_emp} resembles L2 regularization in non-Bayesian training, but we are penalizing variational parameters (\textbf{M}, \textbf{m}) and learning the \emph{distribution} over $\Theta=(\textbf{W}, \textbf{b})$. Posterior predictive distributions \citep{gelman2014understanding} can  be constructed though MC draws of $\Theta$ (for details, see Appendix  \ref{sec_uncertainity}). 

We summarize the forward pass for EMP/ECMP in Algorithm \ref{alg:emp_pseudocode}, and showcase an example implemented in PyramidNet using PyTorch in the Supplementary.  We theoretically justify the unbiasedness of the MC integration and therefore the  convergence of Algorithm \ref{alg:emp_pseudocode} in Appendix \ref{sec_convergence}.

\subsection{Choice of Model Patch}
\normalsize 

Following \citet{mudrakarta2018k}, we recommend that model patches $\Theta_{\mathrm{patch}}$ be chosen from parameter-efficient layers that are disproportionately expressive. In most networks, that would be the normalization or affine layers \citep{dumoulin2017learned,ghiasi2017exploring,karras2018stylebased,kim2017dynamic,dumoulin2018feature-wise,perez2017film} (for example, the batch normalization layers are only $0.1\%$ of the parameters in InceptionV3 \citep{szegedy2015going}). It typically also helps to include the encoder/decoder layers, which are the input/output layers in most networks, since they interface with the data.

We can compute the layer-wise overhead as follows, assuming $N$ total parameters in the network and $K$ mixture components in each layer. For \textbf{Batch Normalization (BN)} layers, given a fully connected layer with size $N_1 \times N_2$ followed by a BN Layer, the parameter overhead is $2(K-1) N_2$ (scales $\mathcal{O}(K \sqrt{{N}/{L}})$). Given a convolution layer with $N_1$ input channels, $N_2$ output channels, kernel width $k$, the parameter overhead is $2(K-1) N_2$ (scales $\mathcal{O}(K\sqrt{N/L}/k^2 )$). For \textbf{Input/Output} layers, given a feedforward network with $L$ fully connected layers, where layer $i$ has input dimension $I_i$ and output dimension $O_i$, the parameter overhead is $(K-1)((I_1 +1) O_1 + (I_L +1) O_L)$. Given a feedforward network with $L$ convolutional layers having $I_i$ input channels, $O_i$ output channels, kernel width $k_i$, and no bias for layer $i$, the parameter overhead is $(K-1)(k_1^2 I_1 O_1 + k_L^2 I_L O_L)$. They scale $\mathcal{O}(\frac{KN}{L})$.

\subsection{Trade-off Between Ensemble Expressiveness and Computational Resources}
\label{section:asymptotic}
\begin{table}
\vspace{-1em}
\caption{Comparison of ensemble size and resource requirements between Ensemble-Based VBNNs}
\label{table:asymptotic_mlp}
\centering
\footnotesize
\begin{tabular}{p{2.3cm} p{1.6cm} p{2.5cm} p{1.55cm} p{1.65cm} p{1.7cm} }
 \toprule
 Method & Effective Ensemble Size & Memory Overhead\newline ($K << L,H$) & Parameters in MLP & Parameters in ResNet-18 & Parameters in PyramidNet\\
 \midrule
 EMP & $\mathcal{O}(K^L)$ & $\mathcal{O}(K(H^2 + LH))$ & $1,129,100$ & $13,779,912$ & $28,825,299$\\
 ECMP & $\mathcal{O}(K^{H^2 + LH})$ & $\mathcal{O}(K(H^2 + LH))$ & $1,129,100$ & $13,779,912$ & $28,825,299$\\
 Dropout & $\mathcal{O}(2^{LH})$ & $\mathcal{O}(LH)$ & $1,009,900$ & $11,689,512$ & $28,511,307$\\
 DropConnect & $\mathcal{O}(2^{LH^2})$ & $\mathcal{O}(LH^2)$ & $1,009,900$ & $11,689,512$ & $28,511,307$\\
 Explicit Ensemble & $\mathcal{O}(K)$ & $\mathcal{O}(KLH^2)$ & $5,049,500$ & $58,447,560$ & $142,556,535$\\
 \bottomrule
\end{tabular}
\vspace{-1em}
\end{table}
\normalsize

 \begin{wrapfigure}{r}{0.45\textwidth}
 \vspace{-3em}
  \begin{center}
    \includegraphics[width=\textwidth]{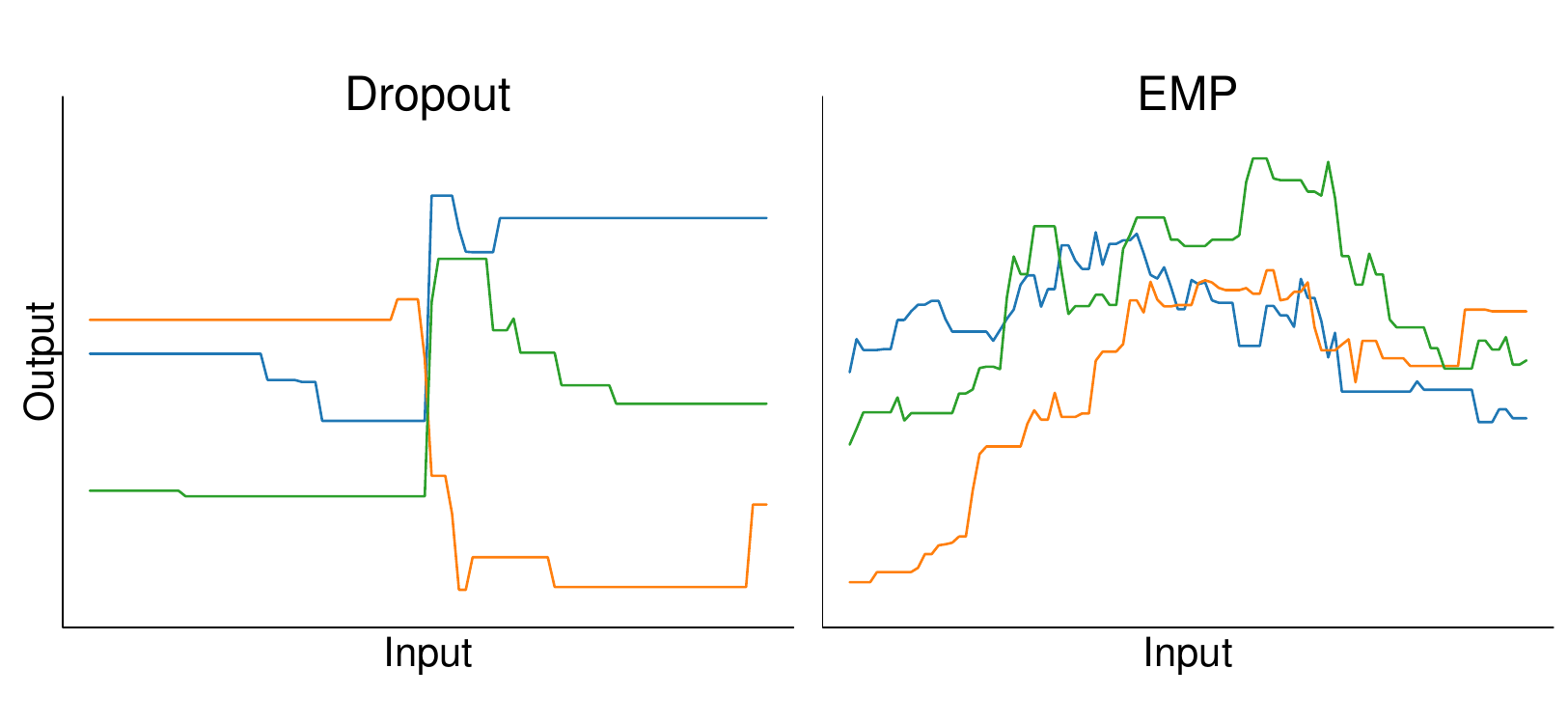}
  \end{center}
  \vspace{-1.5em}
  \caption{\footnotesize We simulate functions from a network with one single hidden layer. The weights are generated from a normal prior via dropout or EMP. EMP expresses finer details. See Appendix \ref{section:toy_simulation}. \vspace{-2em}}\label{fig:prior}
\end{wrapfigure}
In theory, any posterior distribution can be approximated by an infinite mixture of Gaussians, therefore a larger ensemble size should result in a more expressive variational approximation. However, this may incur higher, maybe infeasible, memory overhead. 

In Table \ref{table:asymptotic_mlp}, we compare the effective ensemble size (product of each layer) and resource requirements between ensemble-based VBNN methods, using a feedforward architecture that has $L$ fully connected layers with $H$ hidden units each (total number of parameters $N = \mathcal{O}(LH^2)$). We assume no in-place modification for dropout and DropConnect, as well as patched BN and output layers for EMP/ECMP, with $K$ mixture components for each layer. We assume $K=5$ to compute parameter counts for an MLP ($L=100,H=100$), ResNet-18, and PyramidNet (depth$=110$, $\alpha=270$, no-bottleneck). Memory overhead is dominated by parameter overhead but also includes temporary variables stored during program execution.

A caveat to the ensemble size analysis is that the kinds of networks expressible in different ensembles are different, with explicit ensembles more flexible than EMP/ECMP, which are in turn more flexible than dropout/DropConnect (Figure \ref{fig:prior}). That said, implicit ensembles are the product of ensembles in each layer, making the effective ensemble size exponentially larger than explicit ensembles. 


We observe that ECMP expresses an asymptotically larger ensemble size than dropout, while requiring a small extra memory and parameter overhead, thus being a good trade-off between expressiveness and computational cost. The number of patch layers $|\Lambda_p|$ as well as the number of components in each layer's ensemble $K_i$ can be tuned as hyperparameters in EMP and ECMP, thus allowing the programmer the flexibility of adapting to her available compute budget and desired approximation accuracy. In our experiments, we choose $K=5$ and find it achieves reasonable expressiveness.


\section{Related Work}
\label{section:related_work}
There is a long history of approximate Bayesian inference for neural networks \citep{lampinen2001bayesian}. \citet{mackay1992practical} proposed the use of the Laplace approximation, where L2 regularization can be viewed as a special case. \citet{neal1992bayesian, neal2012bayesian} demonstrated the use of 
Markov chain MC (MCMC) methods,  which are memory intensive because of the need to store samples. There has been recent work that attempts to sidestep this by learning a Generative Adversarial Network \citep{goodfellow2014generative} to recreate these samples \citep{wang2018adversarial}. 

Variational Bayesian neural networks require significantly fewer computational resources. The downside is that variational inference is not guaranteed to reasonably approximate the true posterior \citep{yao2018yes}, especially given the \emph{multi-modal} nature of neural networks \citep{baltruvsaitis2019multimodal}. \citep{graves2011practical}  proposed a factorial Gaussian approximation, and presented a biased estimator for the variational parameters, which \citep{blundell2015weight} subsequently improved with an unbiased estimator and a scale mixture prior. Standard Bernoulli and Gaussian dropout can both be interpreted as variational inference \citep{gal2016dropout,kingma2015variational}. We improve upon \citep{gal2016dropout} by presenting a general variational family for ensemble-based Bayesian neural networks and showing that Bernoulli dropout is a special case. 
Other variational Bayesian neural networks include \citep{louizos2017multiplicative,krueger2017bayesian}, which use a sequence of invertible transformations known as a normalizing flow to increase the expressiveness of the approximate posterior. Normalizing flow methods incur a significant computational and memory overhead. \citep{louizos2016structured} proposed a  parameter-efficient matrix Gaussian approximate posterior by assuming independent rows and columns, which is orthogonal to our work.

Using mixture distributions to enrich the expressiveness of variational Bayes is not a new idea. Earlier work has either used a mixture mean-field approximation to 
model the posterior   \citep{bishop1998approximating,jaakkola1998improving, zobay2014variational, gershman2012nonparametric, miller2017variational} or variational parameters \citep{ranganath2016hierarchical}. The variational family \eqref{eq_variationalfamily} we consider is essentially a mixture mean-field method. However, a direct application of mixture variational methods is prohibitively expensive in large models, where even a non-mixture mean-field approximation incurs a $100\%$ parameter overhead. Our methods, by virtue of a light parameter overhead, are tailored for large Bayesian neural networks. In the proposed methods, we marginalize out the discrete variables by one MC draw in the training step, which resembles particle variational methods \citep{saeedi2017variational}.

Explicit ensembles of neural networks can be used to model uncertainty \citep{lakshminarayanan2017simple}, and even done in parameter-efficient ways \citep{huang2017snapshot,izmailov2018averaging}. However, these methods typically lack a Bayesian interpretation, which is a principled paradigm of modeling uncertainty. Our work addresses this shortcoming. \citep{pearce2018bayesian} proposed an ensemble-based Bayesian neural network. Our variational family is more general, covering implicit ensembles as well as parameter-efficient members like EMP and ECMP.


\section{Experiments}\label{section:experiments}
We investigate our methods using deep residual networks on ImageNet, ImageNet-C, CIFAR-100, and a shallow network on a collection of ten regression datasets. Our aim is to show how our proposed methods can be used to Bayesianize existing deep neural network architectures, rather than show state-of-the-art results. As such, we do not tune hyper-parameters and use Adam \citep{kingma2014adam} on the default settings. We use ensemble size $K=5$ for each layer. While it is not uncommon to report the best test accuracy found during the course of training, we only evaluate the models found at the end of training.
More implementation details can be found in Appendix \ref{section:moreexp}. 
\paragraph{ImageNet}\label{section:imagenet}
ImageNet is a $1000$-class image classification dataset with $1.28M$ training images and $50K$ validation images used for testing \citep{ILSVRC15}. It is a commonly used benchmark in deep learning, but to our knowledge, no Bayesian neural network has been reported on it, likely due to the parameter inefficiency of standard methods. We evaluated our methods on ResNet-18 against these metrics: parameter overhead, top-5/top-1 test accuracy, expected calibration error (ECE), maximum calibration error (MCE), and robustness against common image corruptions (using the ImageNet-C dataset).

Table \ref{table:imagenet} shows that dropout is slightly better calibrated than the vanilla (non-Bayesian) model, but has lower test accuracy. Here, Bayesianizing a network by dropout forces a trade-off between test accuracy and calibration. At the cost of a slight parameter overhead, EMP and ECMP avoid this trade-off by having both higher test accuracy and lower calibration error than the vanilla and dropout models. In particular, ECMP (patched on BN and output layers) achieves \textbf{almost perfect calibration}.

The output layer in ResNet-18 is not parameter-efficient, incurring a $17.6\%$ overhead over just model-patching the BN layers. But it significantly improves the model performance and calibration for EMP and ECMP. For reference, a $6\%$ increase in top-5 accuracy on ImageNet corresponds to approximately $5$ years of progress made by the community \citep{recht2019imagenet}, so the $0.5\%$ improvement in top-5 accuracy for EMP/ECMP (BN+output) over the vanilla model is significant---half a year's progress.

The vanilla model took approximately a week to train on our multi-GPU system, with our Bayesian models requiring only a few additional hours. 
In comparison, related work that incurs a $100\%$ parameter overhead would have required twice as many FLOPs,
and may no longer fit in GPU memory. If so, we would have needed to halve the batch size, which doubles the training time \citep{mccandlish2018empirical}.
Thus, even if we discount the longer convergence time required by a bigger model, the cost of Bayesianizing a neural network via a method with a $100\%$ parameter overhead is a \textbf{2-4x} longer training time.
This  discourages the use of parameter-inefficient VBNN methods on deep learning scale datasets (ImageNet) and architectures (ResNet-18).

\begin{table}
\vspace{-1.0em}
\caption{ResNet-18 on ImageNet}
\label{table:imagenet}
\centering
\footnotesize
\begin{tabular}{p{2.3cm} p{1.4cm} p{1.2cm} p{1.2cm} p{2.2cm} p{2.2cm}}
 \toprule
 Method & Parameter Overhead & Top-5\newline Accuracy & Top-1\newline Accuracy & Expected\newline Calibration Error & Maximum\newline Calibration Error\\
 \midrule
 EMP (BN+out) & $17.9\%$ & $\textbf{87.2\%}$ & $67.0\%$ & $3.91\%$ & $6.83\%$ \\
 EMP (BN) & $0.328\%$ & $86.8\%$ & $66.6\%$ & $5.74\%$ & $11.0\%$ \\
 ECMP (BN+out) & $17.9\%$ & $\textbf{87.2\%}$ & $\textbf{67.1\%}$ & $\textbf{1.65\%}$ & $\textbf{3.15\%}$\\
 ECMP (BN) & $0.328\%$ & $86.8\%$ & $66.4\%$ & $4.62\%$ & $8.07\%$\\
 Dropout & ${0.00\%}$ & $86.7\%$ & $65.9\%$ & $7.61\%$ & $14.0\%$\\
 Vanilla & $0.00\%$ & $86.7\%$ & $66.1\%$ & $8.09\%$ & $14.2\%$\\
 \bottomrule
\end{tabular}
\end{table}
\normalsize


\begin{figure}
\vspace{-1.3em}
\CenterFloatBoxes
\begin{floatrow}
\ffigbox{%
    \includegraphics[width=0.45\textwidth]{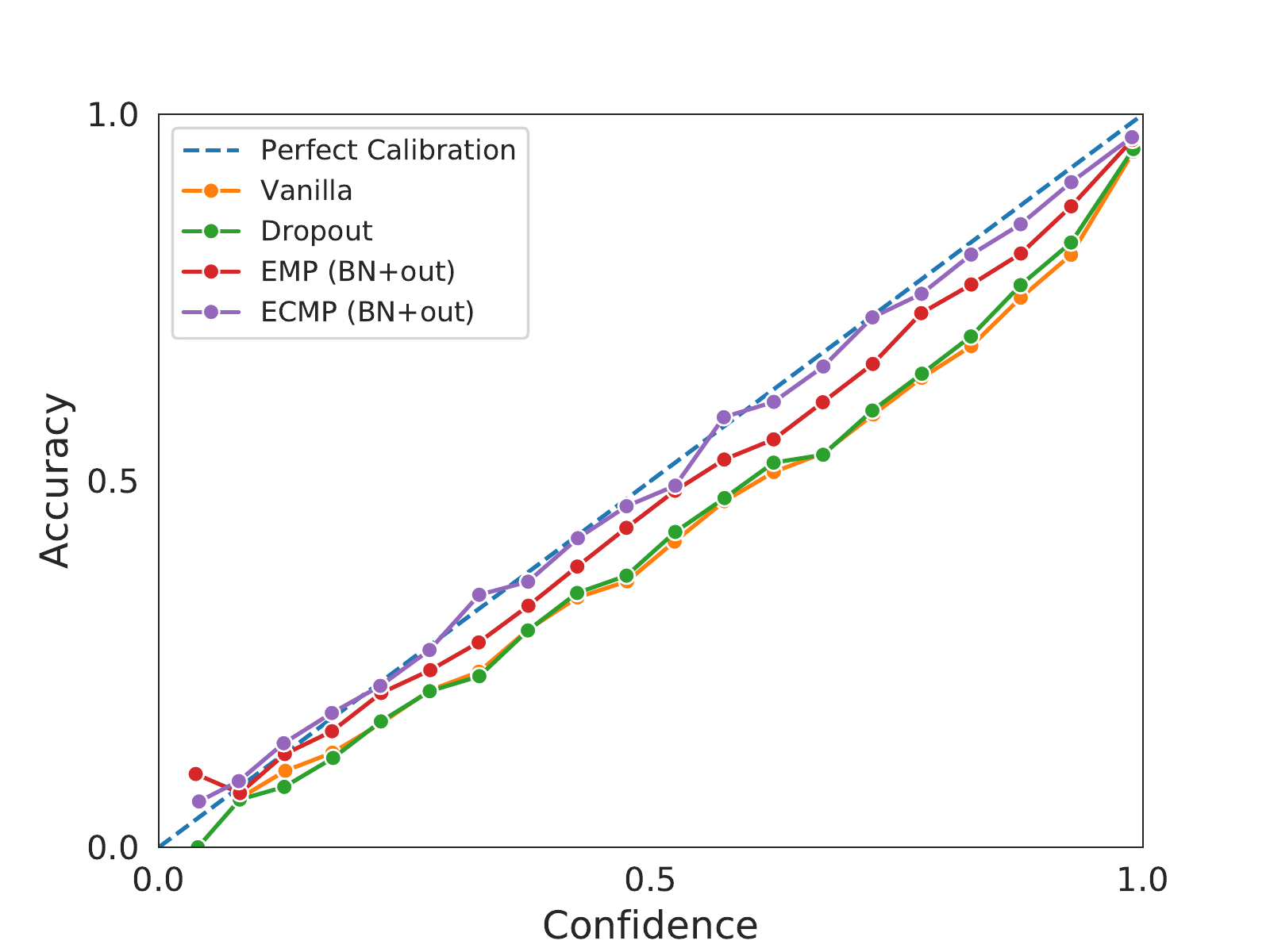}
    \centering
    }{%
    \vspace{-0.5em}
    \caption{\footnotesize The calibration curve for ResNet-18 on ImageNet. EMP and ECMP are better calibrated.}
    \label{fig:imagenet_calibration}
}
\killfloatstyle
\ffigbox{%
    \includegraphics[width=0.45\textwidth]{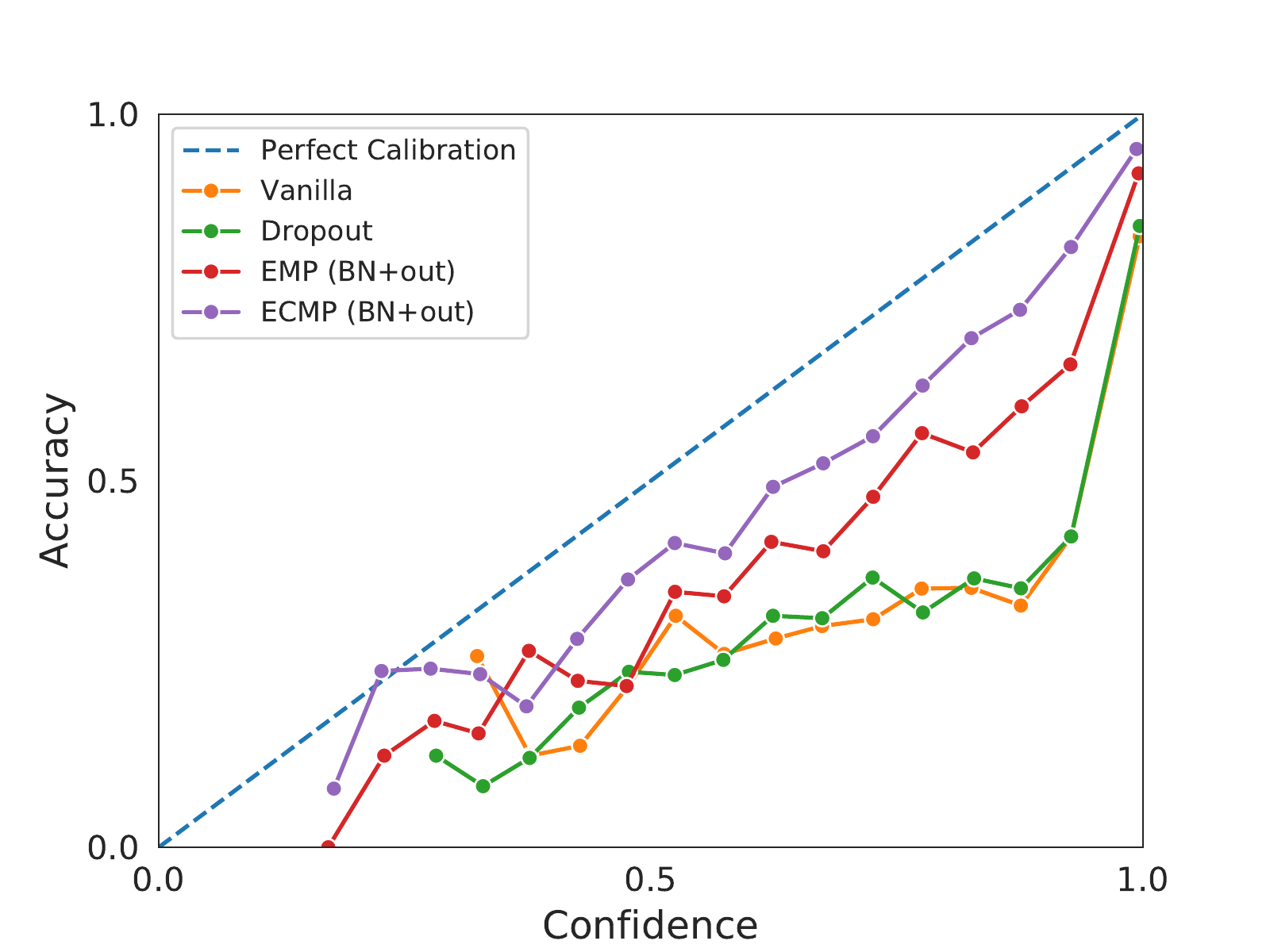}
    \centering
    }{%
    \vspace{-0.5em}
    \caption{\footnotesize The calibration curve for PyramidNet on CIFAR-100. EMP and ECMP are better calibrated.}
    \label{fig:cifar100_calibration}
}{}
\end{floatrow}
\end{figure}

\paragraph{ImageNet-C}
ImageNet-C is a dataset that measures the robustness of ImageNet-trained models to fifteen common kinds of image corruptions reflecting realistic artifacts found across four distinct categories and five different levels of severity. We test the ResNet-18 models trained in the previous section with $S=100$ MC samples, and compute the mean corruption error and the relative mean corruption error in Table \ref{table:imagenet-c_mce_main} and \ref{table:imagenet-c_rmce} respectively. Again, EMP and ECMP outperform dropout and the vanilla model, and provide more robust predictions.

\begin{table}
\caption{\footnotesize Mean Corruption Error On ImageNet-C}
\vspace{-1.0em}
\label{table:imagenet-c_mce_main}
\centering
\tiny
\begin{tabular}{p{1.32cm} p{0.4cm} p{0.3cm} p{0.3cm} p{0.35cm} p{0.3cm} p{0.3cm} p{0.3cm} p{0.35cm} p{0.3cm} p{0.3cm} p{0.3cm} p{0.35cm} p{0.3cm} p{0.3cm} p{0.3cm} p{0.3cm} }
 \toprule
 & & \multicolumn{3}{c}{ Noise} & \multicolumn{4}{c}{{ Blur}} & \multicolumn{4}{c}{{ Weather}} & \multicolumn{4}{c}{{ Digital}}\\
 {\fontsize{8}{10}\selectfont Method} & \textbf{mCE} & Gauss & Shot & Impulse & Defoc & Glass & Motion & Zoom & Snow & Frost & Fog & Bright & Cont & Elastic & Pixel & JPEG\\
 \midrule
 EMP (BN+O) & $\textbf{97.8}$ & $\textbf{98.4}$ & $\textbf{98.2}$ & $\textbf{98.5}$ & $\textbf{97.9}$ & $98.8$ & $\textbf{96.9}$ & $\textbf{98.3}$ & $\textbf{99.6}$ & $101.1$ & $\textbf{99.8}$ & $\textbf{97.9}$ & $\textbf{100}$ & $\textbf{95.4}$ & $91.3$ & $\textbf{94.2}$ \\
 ECMP (BN+O) & $99.5$ & $101$ & $101$ & $102$ & $99.5$ & $99.3$ & $97.7$ & $97.8$ & $98.0$ & $\textbf{99.1}$ & $103$ & $99.1$ & $102$ & $97.2$ & $97.3$ & $98.6$\\
 Dropout & $100$ & $99.8$ & $99.4$ & $98.7$ & $101$ & $\textbf{98.5}$ & $100$ & $101$ & $101$ & $101$ & $104$ & $105$ & $102$ & $99.5$ & $97.6$ & $98.5$ \\
 Vanilla & $100$ & $100$ & $100$ & $100$ & $100$ & $100$ & $100$ & $100$ & $100$ & $100$ & $100$ & $100$ & $\textbf{100}$ & $100$ & $100$ & $100$\\
 \bottomrule
\end{tabular}
\end{table}

\begin{figure}
\vspace{-1.0em}
\CenterFloatBoxes
\footnotesize
\begin{floatrow}

\killfloatstyle
\capbtabbox[.5\textwidth]{%
\vspace{-1.0em}

\centering
\footnotesize
\begin{tabular}{p{1.1cm} p{.74cm} p{.7cm} p{.7cm} p{0.7cm} p{0.7cm}}
 \toprule
 Method &Over-\newline head& Top5\newline Acc. & Top1 \newline  Acc. & ECE & MCE\\
 \midrule
 EMP & $1.10\%$ & $\textbf{92.0\%}$ & $\textbf{74.0\%}$ & $13.6\%$ & $28.9\%$ \\
 ECMP & $1.10\%$ & $91.9\%$ & $73.5\%$ & $\textbf{8.73\%}$ & $\textbf{18.1\%}$ \\
 Dropout & $0.00\%$ & $91.4\%$ & $72.0\%$ & $21.1\%$ & $52.3\%$ \\
 Vanilla & $0.00\%$ & $90.9\%$ & $72.6\%$ & $21.6\%$ & $54.6\%$ \\
 \bottomrule
\end{tabular}
\normalsize
}{%
  \caption{PyramidNet on CIFAR-100}
  \label{table:cifar100}\vspace{-0.8em}
}

\ffigbox[.5\textwidth]{
    \includegraphics[width=0.4\textwidth]{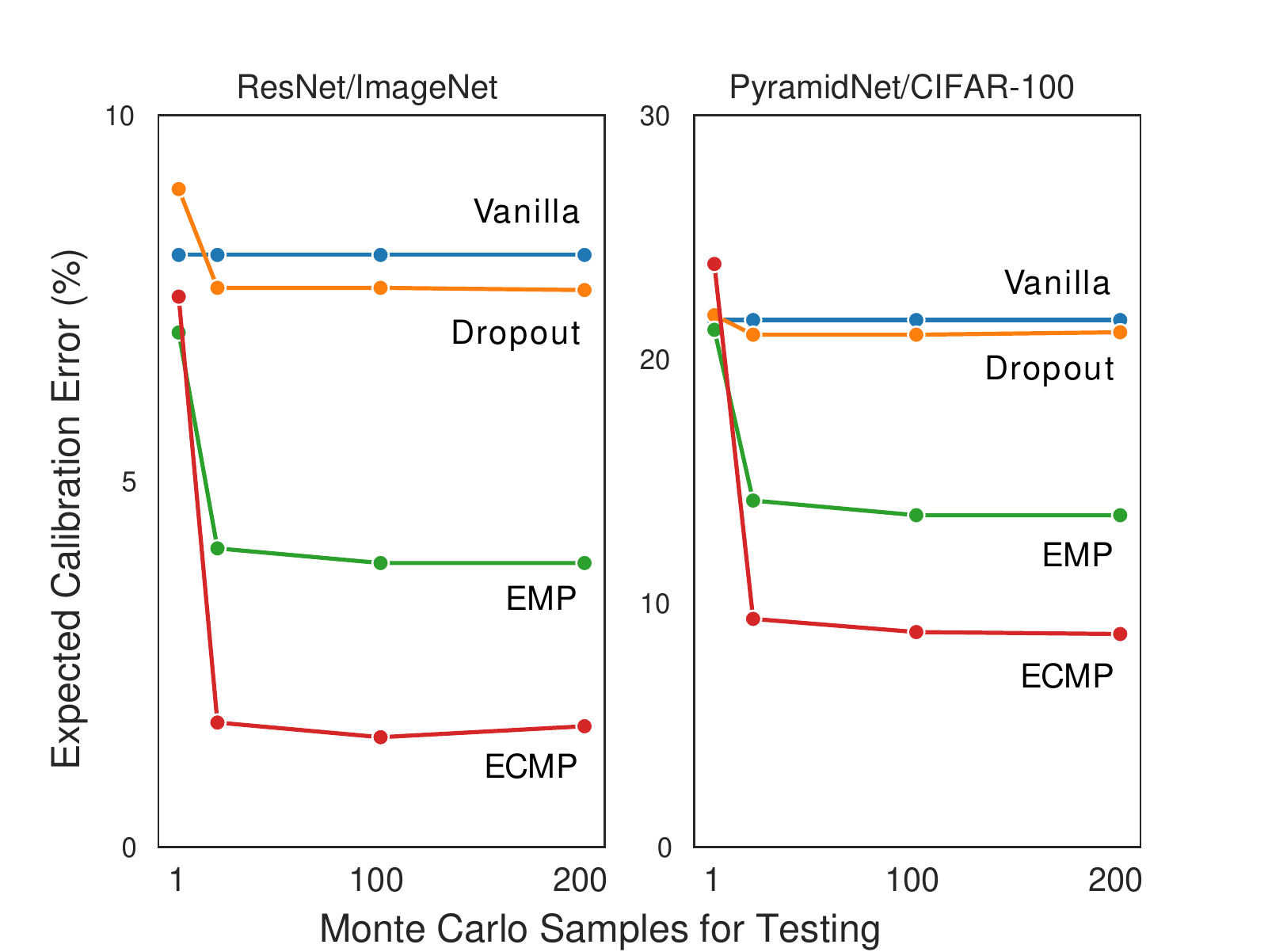}
    \centering
    }{%
    \vspace{-1.0em}
    \caption{\footnotesize How MC sample size affects calibration.
    }
    \label{fig:sample_size_graph}
}
\end{floatrow}
\vspace{-0.5cm}

\end{figure}

\paragraph{CIFAR-100}
\label{section:cifar100}
CIFAR-100 is a $100$-class image classification problem with $500$ training images and $100$ testing images for each class. We test our methods on PyramidNet against top-5/top-1 accuracy and expected/maximum calibration error, as summarized in Table \ref{table:cifar100}. We observe that EMP and ECMP are both better calibrated and more accurate than dropout and the vanilla model. The PyramidNet architecture and the smaller size of CIFAR-100 images contribute to a compact output layer. For a $1.10\%$ parameter overhead, we more than halved the calibration error with ECMP.

We evaluate our methods for ImageNet and CIFAR-100 using $S=200$ MC samples in testing. Each sample potentially captures one mode of the posterior, therefore a higher number of samples results in better calibration, as shown in Figure \ref{fig:sample_size_graph}. This explains why the larger effective ensemble size in our methods is highly desirable for approximating the posterior and calibrating for uncertainty.
  
\normalsize


\paragraph{Regression Experiments}
Following \citep{hernandez2015probabilistic,gal2016dropout,louizos2016structured,lakshminarayanan2017simple}, we test the predictive performance of our methods on a collection of ten regression datasets. We observe that EMP and ECMP consistently outperform dropout and the vanilla model measured by either test root mean squared error or expected log predictive density. Detailed comparisons can be found in Table \ref{table:ten_regression_1} and \ref{table:ten_regression_2} at Appendix \ref{section:ten_regression}.


\section{Discussion and Future Work}
\label{section:conclusion}
Our work bridges implicit and explicit ensembles with a general variational distribution. We focused on scaling VBNNs for deep learning practitioners, and hence proposed two members of the family, EMP and ECMP, that economize both parameter and programming overhead. While common methods like mean-field variational inference double parameter use, making them infeasible for state-of-the-art architectures, our methods scale easily to deep learning scale datasets like ImageNet and architectures like ResNet and PyramidNet. We showed experimentally that VBNNs constructed with Ensemble Model Patching work well with batch-normalized networks, achieving better prediction accuracy and probability calibration than dropout and the non-Bayesian alternative. We hope this work will draw more attention to computationally efficient methods in large scale Bayesian inference.

There are several research directions for future work. First, instead of fixing them to be uniform, the mixing probabilities can be made learnable with an ancestral sampling technique \citep{graves2016stochastic} or post-inference reweighting   \citep{yao2018using}. Second, we can pursue more complicated variational distributions for the patch layers using methods like normalizing flows. Third, we can investigate other potentially superior variants within the general ensemble-based variational family, for example mixing EMP on the normalization layers and dropout/ECMP on the fully connected layers. These methods increase the complexity of implementation (and thus were avoided in this work), but might prove to be more effective at calibration and accuracy (at the expense of some additional parameter overhead).

\newpage
\subsection*{Acknowledgments}
This research was supported in part by the US Defense Advanced
Research Project Agency (DARPA) Lifelong Learning Machines Program, grant HR0011-18-2-0020. The authors would like to thank Aki Vehtari, Andrew Gelman, and Lampros Flokas for helpful discussion and comments.

\bibliography{icml}

\newpage
\appendix
\section*{Appendix}
\section{Background}
\label{section:background}
We provide a brief introduction to the basic technical background of Bayesian neural networks. The reader familiar with the area may want to skip Section \ref{section:background}.
\subsection{Learning a Point Estimate of a Bayesian Neural Network}
In a regression or classification task, a neural network is given input $\mathbf{X}$ and has to model output $\mathbf{Y} = \mathbf{f}(\mathbf{X}, \Theta)$ using weights and biases $\Theta$ in the network. The size of the dataset is denoted $N$. For example,  in regression  $p(\mathbf{Y} | \mathbf{X}, \Theta)$ is  usually assumed to be Gaussian $\mathcal{N}(\mathbf{f}(\mathbf{X}, \Theta), \tau_{output}^{-1} \mathbf{I})$, with $\mathbf{f}(\mathbf{X}, \Theta)$ modeling the epistemic uncertainty and $\tau_{output}^{-1} \mathbf{I}$ modeling the aleatoric uncertainty \citep{kendall2017uncertainties}.  





The weights in the network can be  learned  via maximum a posteriori (MAP) estimation.
\begin{equation}
\begin{split}
\Theta_\mathrm{MAP} = \argmax_{\Theta} \log p(\Theta | \mathbf{X}, \mathbf{Y})
             = \argmax_{\Theta} \log p(\mathbf{Y} | \mathbf{X}, \Theta) + \log p^{\mathrm{prior}}(\Theta).
\end{split}
\end{equation}
The prior on weights $p(\Theta)$ is commonly chosen to be Gaussian, which results in L2 regularization. With a Laplace prior, we end up with L1 regularization instead.

\subsection{Variational Bayesian Neural Networks}
When $\Theta$ is a point estimate, there can be no epistemic uncertainty in $\mathbf{f}(\mathbf{x}, \Theta)$. One of the aims of a Bayesian neural network is to learn the posterior distribution over $\Theta$ to model the epistemic uncertainty in the network.

Variational inference  approximates the posterior distribution by $q(\Theta)$ such that the Kullback-Leibler (KL) divergence between the two, $\mathrm{KL}(q(\Theta)\ ||\ p(\Theta | \mathbf{X}, \mathbf{Y}))$, is minimized. Minimizing the $\mathrm{KL}$ divergence is equivalent to maximizing the log evidence lower bound (ELBO):
\begin{equation}
\label{eqn:elbo}
\mathrm{ELBO}= \int q(\Theta) \log p(\mathbf{Y} | \mathbf{X}, \Theta) \diff \Theta - \mathrm{KL}(q(\Theta)\ ||\ p^{\mathrm{prior}}(\Theta)).
\end{equation}

\subsection{MC dropout}
MC dropout \citep{gal2015dropout,gal2016dropout} interprets dropout \citep{srivastava2014dropout} as variational Bayesian inference in a deep probabilistic model, specifically a deep Gaussian process. It employs a variational distribution factorized over the weights, where each weight factor $\mathbf{W}_i$ is a mixture of Gaussians and each bias factor $\mathbf{b}_i$ is a multivariate Gaussian.

For a network with $L$ dropout layers, each containing $H_i$ hidden units for $i \in [1,L]$ ($H_0$ is the number of input units), we can describe the variational approximation as follows.

\begin{equation}
\begin{split}
\mathbf{W}_{i}&=\mathbf{Z}_{i} \odot (\mathbf{M}_{i}+{\sigma} \mathbf{\epsilon}_{i})+(\mathbf{1}_{H_{i-1} \times H_i}-\mathbf{Z}_{i}) \odot {\sigma} \mathbf{\epsilon}_{i}\\
\mathbf{Z}_{i, h_1 h_2} &= z_{i h_1}\\
z_{i h_1} &\sim \text{Bernoulli}(p_i)\\
\mathbf{\epsilon}_i &\sim \mathcal{N}(\mathbf{0}, \mathbf{I}_{H_{i-1} \times H_i})\\
\mathbf{b}_j&=\mathbf{m}_j+{\sigma} \mathbf{\epsilon}_j\\
\mathbf{\epsilon}_j &\sim \mathcal{N}(\mathbf{0}, \mathbf{I}_{H_j})\\
\text{for}\ \sigma > 0, h_1 &\in [1, H_{i-1}], h_2 \in [1, H_{i}], i \in [1,L], j \in [1,L]\\
\end{split}
\end{equation}

$\mathbf{1}_{H_{i-1} \times H_i}$ denotes a matrix where all the entries are ones. $\odot$ denotes the Hadamard product. $\mathbf{Z}_{i, h_1 h_2}$ denotes the entry at the $h_1$th row and $h_2$th column of the matrix $\mathbf{Z}_i$.

The variational parameters are the Bernoulli probabilities $\{p_i\}_{i=1}^{L}$ and the neural network weights $\{\mathbf{M}_i\}_{i=1}^{L}$, $\{\mathbf{m}_j\}^{L}_{j=1}$. The Bernoulli probabilities are typically fixed and not learned.

\section{Computation of ELBO for the General Ensemble-Based Variational Family}
\label{section:klmog}

The ELBO  can  always be decomposed into the expected log likelihood and the KL divergence between the approximate posterior $q(\Theta)$ and prior $p^{\mathrm{prior}}(\Theta)$
\begin{equation}
\mathrm{ELBO}= \int q(\Theta) \log p(\mathbf{Y} | \mathbf{X}, \Theta) \diff \Theta - \mathrm{KL}(q(\Theta)\ ||\ p^{\mathrm{prior}}(\Theta)).
\end{equation}
The likelihood term, $\int q(\Theta) \log P(\mathbf{Y} | \mathbf{X}, \Theta) \diff \Theta$ can be approximated by MC draws $\Theta_1\dots \Theta_S$ .
$$\int q(\Theta) \log p(\mathbf{Y} | \mathbf{X}, \Theta) \diff \Theta \approx \sum_{s=1}^S  \log p(\mathbf{Y} | \mathbf{X}, \Theta_s).    $$
In particular, for regression problems when assuming a fixed output precision $\tau_{\mathrm{output}}$, 
$$\mathbf{Y}|\Theta, \mathbf{X} \sim \mathcal{N} (\mathbf{Y}|\mathbf{\hat{Y}}, \tau_\mathrm{output}^{-1} \mathbf{I}   ), $$
where $\mathbf{\hat{Y}}$ is the (point estimation) prediction  of $\mathbf{Y}$ in the neural network with input $\mathbf{X}$ and parameter $\Theta$, we have 
$$\int q(\Theta) \log p(\mathbf{Y} | \mathbf{X}, \Theta) \diff \Theta \approx - \frac{\tau_\mathrm{output}}{2S}  \sum_{s=1}^S  ||\mathbf{Y}- \mathbf{\hat {Y}}_s||^2_2.    $$
where $\mathbf{\hat {Y}}_s$ is the (point estimation) prediction  of $\mathbf{Y}$ in the neural network with input $\mathbf{X}$ and parameter $\Theta_s$ for MC draw $s$.

For classification, $$\int q(\Theta) \log p(\mathbf{Y} | \mathbf{X}, \Theta) \diff \Theta  
\approx  \frac{1}{S} \sum_{n=1}^{N} \sum_{s=1}^{S} \log \hat p_{y_{n},s}.
$$ 
where $\hat p_{y_{n},s}$ is the point estimation of $y_n$ in the neural network with the input $X_n$ and the $s$-th MC draw of the parameters $\Theta_s$.

In the following two sections \ref{section:entropy_approximation} and \ref{section:elbo_derivation}, we will prove
\begin{theorem}\label{thm:kl}
In Equation \eqref{eqn:kl_emp}, the KL divergence can be approximated by 
\begin{equation}
\begin{split}
\mathrm{KL}(q(\Theta \mid \mathbf{M}, \mathbf{m})||\ p^{\mathrm{prior}}(\Theta))  &\approx \sum_{i\in \Lambda_p} \sum_{k=1}^{K_i} \frac{\tau_i p_{ik}}{2} || \mathbf{M}_{ik} ||^2_2 + \sum_{j \in \Lambda_p} \sum_{k=1}^{K_j} \frac{\rho_j p_{jk}}{2} || \mathbf{m}_{jk} ||^2_2 \\
&+  \sum_{i\in \Lambda_s}\! \frac{\tau_i}{2} || \mathbf{M}_{i} ||^2_2 + \sum_{j \in \Lambda_p}\!\!  \frac{\rho_j }{2} || \mathbf{m}_{j} ||^2_2 + \mathrm{Constant}, 
\end{split}
\end{equation}
where the constant is referred to as constant with respect to $\mathbf{M}, \mathbf{m}$.
\end{theorem}

\subsection{Approximating the Entropy of a Gaussian Mixture by the Sum of Individual  Entropies}
\label{section:entropy_approximation}
Consider the most general case where the variational distribution $q(\Theta)$ is a mixture of $D$-dimensional Gaussians. $q$ is fully parameterized by $\mu$ and $\Sigma$,
$$q(\Theta \mid \mu,\Sigma )=\sum_{k=1}^{K} p_{k} \mathcal{N}\left(\Theta ; \mathbf{\mu}_{k}, \mathbf{\Sigma}_{k}\right), \quad \Theta \in \mathbb{R}^{D}.$$
where $K$ is the number of components, $\mathbf{\mu}_{i} \in \mathbb{R}^{D} , \Sigma_{k} \in \mathbb{R}^{D \times D}   $ are variational parameters, and $p_k$ is fixed.




 For a mixture of Gaussians, there is no closed form expression of the entropy term $\mathcal{H}(q(\Theta))=-\int q(\Theta) \log q(\Theta) \mathrm{d} \Theta.$ Nevertheless, it can be upper-bounded by the sum of entropies belonging to each individual component. More precisely,
\begin{equation}\label{eq_entropy}
\begin{split}
\mathcal{H}(q(\Theta))
&\leq \sum_{k=1}^{K} p_k\mathcal{H}(\mathcal{N}\left(\Theta ; \mathbf{\mu}_{k}, \mathbf{\Sigma}_{k}\right))\\
&= \sum_{k=1}^{K} \frac{p_{k}}{2}\left(\log \left|\mathbf{\Sigma}_{k}\right|\right) +\frac{KD}{2}(1+\log 2 \pi).
\end{split}
\end{equation} 
The first line is due to the fact that overlap among mixture components reduces the entropy (for proof, see for example, \citet{zobay2014variational}). When dimension $D$ is high, and the number of mixture components $K$ is not large, the overlap among components is negligible. Therefore, we can approximate the entropy using just the second line of \eqref{eq_entropy}. It is a similar approximation to \citet{gal2015dropout}.  



Further, when $p^{\mathrm{prior}}(\Theta)$ is a multivariate normal centered at 0,
$$p^{\mathrm{prior}}= \mathcal{N}\left(\Theta_i | \mathbf{0}, \tau^{-1} \mathbf{I}_{D}\right),$$
the cross-entropy can be computed as 
 \begin{equation}
\begin{split}
  -\int q(\Theta) \log p(\Theta) \mathrm{d} \Theta &=- \sum_{k=1}^{K} p_{k} \int \mathcal{N}\left(\Theta ; \mathbf{\mu}_{k}, \mathbf{\Sigma}_{k}\right) \log \mathcal{N}\left(\mathbf{0}, \tau^{-1} \mathbf{I}_{D}\right) \mathrm{d} \Theta.\\
 &= \frac{D}{2} (\log (2 \pi) - \log \tau) +  \sum_{i=1}^{K}  \frac{\tau p_{i}}{2} ( \mathbf{\mu}_{i}^{T} \mathbf{\mu}_{i} + \operatorname{Tr}\left(\mathbf{\Sigma}_i\right)).
 \end{split}
 \end{equation}

Putting them together, we get:

 \begin{equation}
\mathrm{KL}(q(\Theta)\ ||\ p^{\mathrm{prior}}(\Theta))  \approx  \sum_{i=1}^{K} \frac{p_{k}}{2}( \tau (\mathbf{\mu}_{k}^{T} \mathbf{\mu}_{k}+\operatorname{tr}\left(\mathbf{ \Sigma}_{k}\right)) -\log \left|\mathbf{\Sigma}_{k}\right|)+ \mathrm{Constant}.
 \end{equation}

In many cases for computational simplicity, we set $\mathbf{\Sigma}_{k}=\sigma^2 \mathbf{I}_D$ where $\sigma$ is a fixed constant. Hence, we get:
\begin{equation}
\mathrm{KL}(q(\Theta)\ ||\ p^{\mathrm{prior}}(\Theta)) \approx \sum_{k=1}^{K} \frac{\tau p_{k}}{2} || \mathbf{\mu}_{k} ||^2_2 + \mathrm{Constant}.
\label{section:klmog_eqn}
 \end{equation}

\subsection {Deriving the ELBO for the General Variational Ensemble}
\label{section:elbo_derivation}
Now if we write the variational distribution for all batch parameters to be 
\begin{equation}
\begin{split}
q(\mathbf{W}, \mathbf{b})&= \prod_{i=1}^{L} q(\mathbf{W}_i)  \prod_{j=1}^{L}q(\mathbf{b}_j).\\
\end{split}
\end{equation}

Then for each dimension,   $$q(\mathbf{W}_i)= \sum_k p_{ik} \mathcal{N}\left(\mathbf{W}_i | \mathbf{M}_{ik}, \sigma^2  \mathbf{I}\right), \quad  q(\mathbf{b}_j) = \sum_k p_{jk} \mathcal{N}\left(\mathbf{b}_j ; \mathbf{m}_{jk}, \sigma^2 \mathbf{I}\right). $$

In Equation \eqref{section:klmog_eqn} with $\mu_k=\mathbf{M}_{ik}$ and $\tau=\tau_i$, we obtain
$$\mathrm{KL}( q(\mathbf{W}_i)) ||  p^{\mathrm{prior}}(\mathbf{W}_i)
=\sum_{k=1}^{K_i} \frac{\tau_i p_{ik}}{2} || \mathbf{M}_{ik} ||^2_2 +\mathrm{Constant}. $$
Similarly, 
$$\mathrm{KL}( q(\mathbf{b}_j)) ||  p^{\mathrm{prior}}(\mathbf{b}_j)
= \sum_{k=1}^{K_j} \frac{\rho_j p_{jk}}{2} || \mathbf{m}_{jk} ||^2_2  +\mathrm{Constant}. $$

Finally, each layer is modeled independently in the variational approximation. Thus we get Equation \eqref{eqn:kl_general}:
\begin{equation}
\mathrm{KL}(q(\Theta)||\ p^{\mathrm{prior}}(\Theta)) \approx \sum_{i=1}^{L} \sum_{k=1}^{K_i} \frac{\tau_i p_{ik}}{2} || \mathbf{M}_{ik} ||^2_2 + \sum_{j=1}^{L} \sum_{k=1}^{K_j} \frac{\rho_j p_{jk}}{2} || \mathbf{m}_{jk} ||^2_2 + \mathrm{Constant}.
\end{equation}

In the presence of model patching, 
$$q(\Theta_{\mathrm{shared}})= \prod_{i \in \Lambda_s  } q(\textbf{W}_{i} ) q(\textbf{b}_{i})  = \prod_{i \in \Lambda_s  } \mathcal{N}(\textbf{W}_{i}| \textbf{M}_i, \sigma_s \textbf{I})   \mathcal{N}(\textbf{b}_{i}| \textbf{m}_i, \sigma_s \textbf{I}).$$
Then for $i \in \Lambda_s$, the KL term is just the KL divergence between two mean-field Gaussians:
$$\mathrm{KL}( q(\textbf{W}_{i} ) || \mathcal{N}( \mathbf{W}_i| \mathbf{0}, \tau_i^{-1} \mathbf{I}))= \frac{\tau_i  }{2} || \mathbf{M}_{ik} ||^2_2 +C, \forall i \in \Lambda_s,$$
$$\mathrm{KL}( q(\textbf{b}_{i} ) || \mathcal{N}( \mathbf{b}_i| \mathbf{0}, \rho_j^{-1} \mathbf{I}))= \frac{\rho_j  }{2} || \mathbf{m}_{i} ||^2_2 +C,  \forall i \in \Lambda_s.$$

This leads to the following
\begin{equation}
\begin{split}
\mathrm{KL}(q(\Theta)||\ p^{\mathrm{prior}}(\Theta))  &\approx \sum_{i\in \Lambda_p} \sum_{k=1}^{K_i} \frac{\tau_i p_{ik}}{2} || \mathbf{M}_{ik} ||^2_2 + \sum_{j \in \Lambda_p} \sum_{k=1}^{K_j} \frac{\rho_j p_{jk}}{2} || \mathbf{m}_{jk} ||^2_2 \\
&+  \sum_{i\in \Lambda_s}\! \frac{\tau_i}{2} || \mathbf{M}_{i} ||^2_2 + \sum_{j \in \Lambda_p}\!\!  \frac{\rho_j }{2} || \mathbf{m}_{j} ||^2_2 + \mathrm{Constant}. 
\end{split}
\end{equation}
which is precisely Equation \eqref{eqn:kl_emp}. To arrive at the ELBO, we just have to combine the $\mathrm{KL}$ term and the likelihood term.
For regression problems, this becomes

\begin{equation}
\begin{split}\label{eq_elbo_reg}
-\mathrm{ELBO}  &=  \frac{\tau_\mathrm{output}}{2S}  \sum_{s=1}^S  \sum_{n=1}^{N}  ||\mathbf{Y_n}-\hat{Y}_{n,s}||^2_2  \\
&+  \sum_{i\in \Lambda_p} \sum_{k=1}^{K_i} \frac{\tau_i p_{ik}}{2} || \mathbf{M}_{ik} ||^2_2 + \sum_{j \in \Lambda_p} \sum_{k=1}^{K_j} \frac{\rho_j p_{jk}}{2} || \mathbf{m}_{jk} ||^2_2 \\
&+  \sum_{i\in \Lambda_s}\! \frac{\tau_i}{2} || \mathbf{M}_{i} ||^2_2 + \sum_{j \in \Lambda_p}\!\!  \frac{\rho_j }{2} || \mathbf{m}_{j} ||^2_2. 
\end{split}
\end{equation}

For classification, this becomes

\begin{equation}
\begin{split}
-\mathrm{ELBO}  &=   -\frac{1}{S}   \sum_{s=1}^S \sum_{n=1}^{N}  \log \hat p_{y_{n},s}\\
&+  \sum_{i\in \Lambda_p} \sum_{k=1}^{K_i} \frac{\tau_i p_{ik}}{2} || \mathbf{M}_{ik} ||^2_2 + \sum_{j \in \Lambda_p} \sum_{k=1}^{K_j} \frac{\rho_j p_{jk}}{2} || \mathbf{m}_{jk} ||^2_2 \\
&+  \sum_{i\in \Lambda_s}\! \frac{\tau_i}{2} || \mathbf{M}_{i} ||^2_2 + \sum_{j \in \Lambda_p}\!\!  \frac{\rho_j }{2} || \mathbf{m}_{j} ||^2_2. 
\end{split}
\end{equation}

\subsection{Posterior Uncertainty} \label{sec_uncertainity}
In training, we run stochastic gradient descent (SGD) with one MC draw ($S$=1) for gradient evaluation. Hence, maximizing the ELBO above is going to resemble the non-Bayesian training that minimizes the usual squared error or cross entropy plus the L2 regularization. This interpretation makes it easy for deep learning programmers to incorporate Bayesian training into their neural networks without much additional programming overhead. However, we emphasize the fundamental differences between conventional point estimation with  L2 regularization and our Bayesian approach:
\begin{itemize}
    \item The L2 regularization is over all variational parameters \textbf{M} and \textbf{m}, not over neural net weights \textbf{W} and \textbf{b}.
    \item Even if we run SGD with one MC draw with one realization of the categorical variable $Z$ at each iteration, the MC gradient is still unbiased. Thus, the optimization converges to the desired variational distribution.
    \item In the testing phase, we will draw $S>1$ to obtain the approximate posterior distribution $\Theta_1, \dots, \Theta_S$.
\end{itemize}
In particular,  we are able to obtain the posterior predictive distribution for the whole model using the variational approximation. The posterior predictive density at a new input $x^*$ can be approximated by 
\begin{equation}
\begin{split}
p(\mathbf{y^*}|\mathbf{x^*},\mathbf{X},\mathbf{Y}) = \int p(\mathbf{y^*}|\mathbf{x^*},\Theta)\ p(\Theta|\mathbf{X},\mathbf{Y}) \diff \Theta \approx 1/S \sum_{s=1}^S \int p(\mathbf{y^*}|\mathbf{x^*},\Theta_s).\ 
\end{split}
\end{equation} 

Any posterior predictive check and posterior uncertainty can then be performed through samping $\{\Theta_s\}$ and then $\{y^*\}$ from $p(\cdot|\mathbf{x^*},\Theta_s)$.

Denote $\bs f ( \mathbf{x^*}, \Theta )$ to be the prediction of outcome $y^*$ at input ${x^*}$. Then the predictive mean and variance can be calculated through MC estimation
\begin{equation}
\begin{split}
\expect^{\mathrm{post}} [y^*| \mathbf{x^*}] &approx
\int {q(\Theta)} \bs{f}(\mathbf{x^*}, \Theta) \diff \Theta \approx \frac{1}{S} \sum_{s=1}^S \bs{f}(\mathbf{x^*}, \Theta_s),\\
\text{Var}^{\mathrm{post}} [y^*| \mathbf{x^*}  ] &\approx \ \frac{1}{S} \sum_{s=1}^S \bs{f}(\bs{x^*}, \Theta_s)^T \bs{f}(\bs{x^*}, \Theta_s) - \big( \expect^{\mathrm{post}} [y^*| \mathbf{x^*}]\big)^T   \expect^{\mathrm{post}}[ y^*| \mathbf{x^*}] + \tau_{output}^{-1} \bs{I}.\\
\end{split}
\end{equation}

\subsection{Stochastic Gradients, MC Integration, and Marginalization of Discrete Variables}\label{sec_convergence}
Theorem \ref{thm:kl} establishes a closed form approximation of KL divergence in the ELBO. What remains left is the expected log likelihood, which is typically estimated through MC integration.  We justify the use of Algorithm \ref{alg:emp_pseudocode} with the following theorem.

\begin{theorem}
The gradient evaluation in Algorithm \ref{alg:emp_pseudocode} is unbiased, and thus SGD will converge to its (local) optimum given other regularization conditions.
\end{theorem}

Algorithm \ref{alg:emp_pseudocode} is implemented through SGD. The entropy term has a closed form.  Essentially, we are using MC estimation three times for the log likelihood term:
\begin{itemize}
    \item We use a minibatch.
    \item We draw MC sample $\theta$ from  $q(\Theta| \mathbf{M,m})$ (by convention, we use  $\theta$  to emphasize that it is one MC realization of  $\Theta$.
    The log likelihhod and its gradient can be evaluated through the  equation   
    \begin{equation}
\mathrm{E}_q\log p(y|\Theta, x)= \int q(\Theta) \log p(y|\Theta, x) \diff \Theta \approx \log p(y|\theta, x).  
    \end{equation}
    \item Indeed, we do not have to derive the explicit form for $q(\Theta|\mathbf{M,m})$, as it depends on discrete variables $\mathbf{Z}$ (integers that indicate the assignments of mixture components). However, we draw one realization of  the discrete assignment $z$ for each layer (again, we use $z$ to emphasize it is one realization of $\mathbf{Z}$). That approximates
    $$ q(\Theta|\mathbf{M,m}) = \int q(\Theta|\mathbf{M,m},Z) q(\mathbf{Z}) \diff Z \approx q(\Theta|\mathbf{M,m}, z).  $$ 
    where $q(\Theta|\mathbf{M,m}, \mathbf{Z})$ is from the construction in Equation \eqref{eq_variationalfamily},
    and $q(\mathbf{Z})$ is a multinoulli distribution specified from before. 
 \end{itemize}
 In all these three steps, the MC approximations are unbiased even with one MC draw, hence so will the gradient of the ELBO. More precisely,  we estimate the likelihood term in the ELBO with the following MC approximation
\begin{equation}\label{eq_unbias}
\begin{split}
& \int q(\Theta |\mathbf{M,m}) \log p(\mathbf{Y} | \mathbf{X}, \Theta) \diff \Theta\\
=&\int  q(\Theta |\mathbf{M,m,Z}) q(\mathbf{Z}) \log p(\mathbf{Y} | \mathbf{X}, \Theta) \diff \Theta\\
\approx&  \log p(\mathbf{Y} | \mathbf{X}, \Theta=\theta),
\end{split}
\end{equation}
where we first draw a realization $z$ from $q(\mathbf{Z})$, and draw a realization $\theta$ from $q(\Theta |\mathbf{M,m}, \mathbf{Z}=z)$, which is exactly what Algorithm \ref{alg:emp_pseudocode} does.  The results are similar where the number of draws $S>1$.

Approximation \eqref{eq_unbias} is always unbiased, based on which the reparametrized gradients $$ \frac{\partial}{\partial \mathbf{M}} \int q(\Theta |\mathbf{M,m}) \log p(\mathbf{Y} | \mathbf{X}, \Theta) \diff \Theta, \quad  \frac{\partial}{\partial \mathbf{m}} \int q(\Theta |\mathbf{M,m}) \log p(\mathbf{Y} | \mathbf{X}, \Theta) \diff \Theta.$$
are also unbiased.  Therefore the convergence theorem of SGD holds.

It remains unclear how many MC draws ($S$) will be the most efficient for training. In the limit where $S=[\text{Effective Ensemble Size}]$, we can marginalize all the discrete variables $\mathbf{Z}$ and get $q(\theta| \mathbf{M,m})$ exactly.  On the other hand, one MC draw is commonly used in practice, and has been commonly reported to be the most efficient setting in variational inference \citep{kucukelbir2017automatic, yao2018yes}. 

\paragraph{Implicit Variational Distribution} We also emphasize that  $\mathbf{Z}$  in Algorithm 1 is a three way tensor, $\mathbf{Z}=\{\mathbf{Z}_{i,h_1, h_2}\}$, where $i$ indexes the layer, and $h_1$, $h_2$ are the indices of the parameter of the weight matrix in layer $i$.
Marginally,  each element ${\mathbf{Z}_{i,h_1, h_2}}$ is from a multinoulli with its corresponding mixing variable. However, different $\mathbf{Z}_{i,h1, h2}$ are not necessarily \emph{independent}. For example, in EMP, all the $\mathbf{Z}={\mathbf{Z}_{i,h_1, h_2}}$ are equal for the same layer $i$.

Writing down the joint distribution of $q(\mathbf{Z})$ can be messy. Nevertheless, in our MC integration, we are only required to be able to sample $\mathbf{Z}$. In this sense,  we are constructing an implicit variational distribution through different constructions of the assignment $\mathbf{Z}$.

\section{Implementation and Hyper-parameters}
\label{section:parallelization}
\subsection{Tuning Hyper-parameters}
From the Bayesian perspective, the hyper-parameters,  which include the prior precisions $\tau, \rho$,  the mixing probability $\mathbf{p}$, and the output precision  $\tau_{out}$, should also be taken into account when evaluating model uncertainty.

Equation \eqref{eqn:kl_emp} can be either extended or simplified. We can extend it by including all the parameters as variational parameters but this significantly increases both the parameter and programming overhead. To simplify the implementation, we can assume uniformity and rewrite the loss function as L2 regularization. Then, regression \eqref{eq_elbo_reg} becomes  
\begin{equation}
\begin{split}
\mathrm{Loss}  &=  \text{Mean Squared Error}\\
&+ \lambda_1  \sum_{i\in \Lambda_p} \sum_{k=1}^{K_i} || \mathbf{M}_{ik} ||^2_2 +  \lambda_2  \sum_{j \in \Lambda_p} \sum_{k=1}^{K_j}   || \mathbf{m}_{jk} ||^2_2 \\
&+  \lambda_3   \sum_{i\in \Lambda_s}\!   || \mathbf{M}_{i} ||^2_2 +  \lambda_4  \sum_{j \in \Lambda_p} || \mathbf{m}_{j} ||^2_2. 
\end{split}
\end{equation}

Then, only four terms $\lambda_1$, $\lambda_2$, $\lambda_3$, $\lambda_4$ have to be tuned. It is an interesting research question to determine how one can tune these parameters to obtain better calibration and posterior uncertainty. 

In our experiments, we simply use an arbitrarily chosen value

$$\lambda_1=\lambda_2=\lambda_3=\lambda_4=\frac{0.001}{[\text{batch size}]}$$

\subsection{Parallelization}
Since all the training in Ensemble Model Patching is done via SGD and backpropagation, it can be easily trained with a regular distributed SGD algorithm with no modifications, like the one outlined in \citet{mudrakarta2018k}. It is also possible to speed the training process by using Coordinate Descent. The learning of $\Theta_{shared}$ and $\Theta_{patch}$ can be split into two alternating phases, by holding one fixed while the other is being trained. The learning of non-overlapping $\Theta_{patch}$ is trivially lock-free and can be done by separate worker processes. The learning of $\Theta_{shared}$ can be done with distributed SGD as before. This training strategy will be most effective when the ensemble size $K$ is large.

\section{Details and Extensions of the Experiments}
\label{section:moreexp}

\subsection{The Implicit Prior on the Function}
\label{section:toy_simulation}
It is a well-known result that a normal prior on the weights and biases leads to a Gaussian process on the final model $\mathbf{y}=\bs{f}(\mathbf{x})$ when the number of hidden units goes to infinity \citep{neal1996priors,neal1992bayesian,neal2012bayesian}. The variational inference approximation can also be viewed as an implicit prior that restricts the distribution of weights and biases. What prior does it imply on the function $\mathbf{y}=\bs{f}(\mathbf{x})$?

In Figure \ref{fig:prior}, we generate a toy example with one hidden layer: 
$$\mathbf{y}= B+ \sum_{h=1}^H \mathbf{W}_{h}u_h, \qquad u_h= g(b_h+ {w_h}x), \quad h=1,\dots, H $$
We use $(B, W)$ and $(b, w)$ to denote the bias and weights in the output and hidden unit layer.
$g$ is the activation function. We stick to $g(x) = \text{sign}(x)$ because of its theoretical convenience. 

We then generate both $B$ and $b$ from N$(0,20)$, as well as W and w from N$(0,5)$, where the number of hidden units is $H=100$. This approximately results in a Gaussian process.

Now, with probability $0.5$ some weights $W$ and $w$ are dropped to $0$. 
Notice that $x$ will never be expressed in a hidden unit $h$ if $w_h$ is dropped to $0$. Therefore, this implies a rough and piece-wise constant function. The left panel of Figure \ref{fig:prior} simulates three such functions.

By contrast, in EMP, the weight and bias are uniformly  chosen from $K=5$ independent Gaussian components with the same parameter mentioned above. This leads to a smoother function (right panel) that are indeed closer to a Gaussian process and is able to express finer details.

We use this example to demonstrate the restriction of fixing one component to be constant at $0$, which is intrinsic to dropout. Our preliminary experiments involving restricting one of the components in EMP and ECMP to zero also indicate worse performance.

\subsection{Calibration Error}
\label{section:calibration}
The calibration error \citep{guo2017calibration} was computed by first splitting the prediction probability interval $[0,1]$ into $20$ equally sized bins, and then measuring the accuracy, confidence, expected calibration error, and maximum calibration error of the model over these $20$ bins. Let $B_r$ be the set of indices denoting the samples whose prediction probability falls in the interval $(\frac{r-1}{20}, \frac{r}{20}]$ for $r \in [1,20]$. Then we have
\begin{equation}
\begin{split}
\text{accuracy}(B_r) &= \frac{1}{|B_r|} \sum_{i \in B_r} \mathds{1}[\mathbf{\hat{y}}_i = \mathbf{y}_i] \\
\text{confidence}(B_r) &= \frac{1}{|B_r|} \sum_{i \in B_r} \hat{p}_i \\
\text{Expected Calibration Error} &= \sum_{r=1}^{20} \frac{|B_r|}{n} \left| \text{accuracy}(B_r) - \text{confidence}(B_r) \right| \\
&\ \text{where $n$ is the total number of samples} \\
\text{Maximum Calibration Error} &= \max_r \left| \text{accuracy}(B_r) - \text{confidence}(B_r) \right| \\
\end{split}
\end{equation}
If a model predicts a $60\%$ probability that a given sample belongs to a certain class, then it ought to be correct $60\%$ of the time. Intuitively, this means that a perfectly calibrated model should have $\text{confidence}(B_r) = \text{accuracy}(B_r)$.

We remove bins with at most $5$ samples in them to get rid of outliers.

\begin{figure}[ht]
\includegraphics[width=0.5\textwidth]{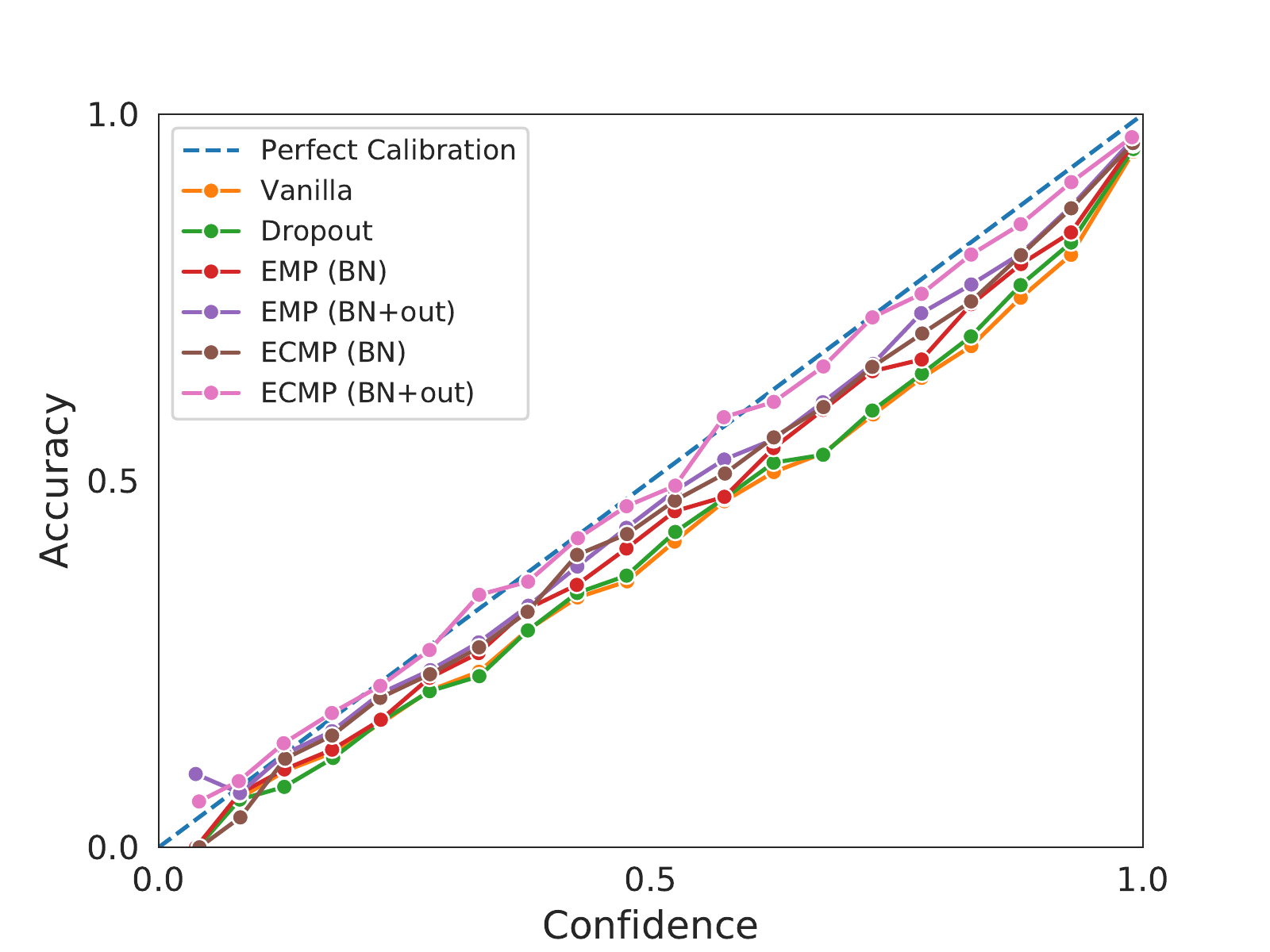}
\centering
\caption{\footnotesize ResNet-18 calibration curves on ImageNet. Patching the output layers in addition to the BN layers improve calibration for both EMP and ECMP. This graph is an expanded version of the graph shown in Figure \ref{fig:imagenet_calibration}.}
\end{figure}
\subsection{ImageNet}
We use the ILSVRC 2012 version of the dataset \citep{ILSVRC15}, as is commonly used to benchmark new architectures in deep learning. As is standard practice, the images are randomly cropped and resized to $224$ by $224$ pixels.

The ResNet-18 was trained for $100$ epochs with batch size $256$ and tested using $S=200$ MC samples. We include different configurations of EMP and ECMP where both the BN and output layers were model patched, and where only the BN layers were patched. This is because the output layer in ResNet-18 is parameter dense due to the size of the images in ImageNet, and we wanted to see the relative effect of including versus excluding the output layer.

We use $p=0.005$ for MC dropout, with the dropout layer occurring before every BN layer. It is difficult to tune the optimal dropout rate without using multiple runs, so this dropout rate is probably not optimal. 

\subsection{ImageNet-C}
\label{section:imagenet-c}
ImageNet-C is a dataset that measures the robustness of ImageNet-trained models to fifteen common kinds of image corruptions reflecting realistic artifacts found across four distinct categories: noise, blur, weather, and digital \citep{hendrycks2019benchmarking}. Each corruption comes in five different levels of severity.

The Mean Corruption Error (mCE) and Relative Mean Corruption Error (rmCE) can be measured as follows:
\begin{equation}
\begin{split}
\text{E}_{c}^{m} &= \text{Top-1 error for model $m$ summed across $5$ different severity levels for corruption $c$,}\\
&\ \ \ \ \ \ \text{where $c=clean$ represents the no-corruption setting}\\
\text{CE}_c^{m} &= \frac{\text{E}_c^{m}}{\text{E}_c^{vanilla}}\\
\text{rCE}_c^{m} &= \frac{\text{E}_c^{m} - \text{E}_{clean}^m}{E_c^{vanilla} - \text{E}_{clean}^m}\\
\text{mCE}^m &= \frac{1}{15} \sum_c \text{CE}_c^m\\
\text{rmCE}^m &= \frac{1}{15} \sum_c \text{rCE}_c^m\\
\end{split}
\end{equation}

Intuitively, the mCE reflects the additional robustness a VBNN method adds to an existing model. But because a model can have lower mCE by virtue of having lower test accuracy in the no-corruption setting. The rmCE taking that into account by measuring the relative change in test performance caused by the corruption.

A priori, we should not expect that being Bayesian will necessarily confer a model with robustness against noise and corruption. For example, we observe that dropout confers no advantage to the vanilla model against corruption.

EMP offers more robustness against common corruptions than ECMP. We hypothesize that this is likely because corruptions introduce more noise at the level of individual weights than at the level of the layer.

\begin{table}[ht]
\caption{\footnotesize Mean Corruption Error in ImageNet-C}
\label{table:imagenet-c_mce}
\centering
\tiny
\begin{tabular}{p{1.32cm} p{0.4cm} p{0.3cm} p{0.3cm} p{0.35cm} p{0.3cm} p{0.3cm} p{0.3cm} p{0.35cm} p{0.3cm} p{0.3cm} p{0.3cm} p{0.35cm} p{0.3cm} p{0.3cm} p{0.3cm} p{0.3cm} }
 \toprule
 & & \multicolumn{3}{c}{ Noise} & \multicolumn{4}{c}{{ Blur}} & \multicolumn{4}{c}{{ Weather}} & \multicolumn{4}{c}{{ Digital}}\\
 {\fontsize{8}{10}\selectfont Method} & \textbf{mCE} & Gauss & Shot & Impulse & Defoc & Glass & Motion & Zoom & Snow & Frost & Fog & Bright & Cont & Elastic & Pixel & JPEG\\
 \midrule
 EMP (BN+O) & $\textbf{97.8}$ & $\textbf{98.4}$ & $\textbf{98.2}$ & $\textbf{98.5}$ & $\textbf{97.9}$ & $98.8$ & $\textbf{96.9}$ & $\textbf{98.3}$ & $\textbf{99.6}$ & $101.1$ & $\textbf{99.8}$ & $\textbf{97.9}$ & $\textbf{100}$ & $\textbf{95.4}$ & $91.3$ & $\textbf{94.2}$ \\
 EMP (BN) & $98.8$ & $99.7$ & $99.6$ & $99.4$ & $\textbf{97.9}$ & $100$ & $98.9$ & $98.4$ & $101$ & $100.9$ & $100$ & $101$ & $\textbf{100}$ & $99.0$ & $\textbf{90.8}$ & $94.7$ \\
 ECMP (BN+O) & $99.5$ & $101$ & $101$ & $102$ & $99.5$ & $99.3$ & $97.7$ & $97.8$ & $98.0$ & $\textbf{99.1}$ & $103$ & $99.1$ & $102$ & $97.2$ & $97.3$ & $98.6$\\
 ECMP (BN) & $99.8$ & $100$ & $99.3$ & $99.9$ & $98.7$ & $99.5$ & $98.3$ & $98.8$ & $101$ & $99.6$ & $103$ & $103$ & $102$ & $99.1$ & $96.0$ & $98.4$ \\
 Dropout & $100$ & $99.8$ & $99.4$ & $98.7$ & $101$ & $\textbf{98.5}$ & $100$ & $101$ & $101$ & $101$ & $104$ & $105$ & $102$ & $99.5$ & $97.6$ & $98.5$ \\
 Vanilla & $100$ & $100$ & $100$ & $100$ & $100$ & $100$ & $100$ & $100$ & $100$ & $100$ & $100$ & $100$ & $\textbf{100}$ & $100$ & $100$ & $100$\\
 \bottomrule
\end{tabular}
\end{table}

\begin{table}[ht]
\caption{\footnotesize Relative Mean Corruption Error  in ImageNet-C}
\label{table:imagenet-c_rmce}
\centering
\tiny
\begin{tabular}{p{1.32cm} p{0.4cm} p{0.3cm} p{0.3cm} p{0.35cm} p{0.3cm} p{0.3cm} p{0.3cm} p{0.35cm} p{0.3cm} p{0.3cm} p{0.3cm} p{0.35cm} p{0.3cm} p{0.3cm} p{0.3cm} p{0.3cm} }
 \toprule
 & & \multicolumn{3}{c}{\fontsize{10}{12}\selectfont Noise} & \multicolumn{4}{c}{{\fontsize{10}{12}\selectfont Blur}} & \multicolumn{4}{c}{{\fontsize{10}{12}\selectfont Weather}} & \multicolumn{4}{c}{{\fontsize{10}{12}\selectfont Digital}}\\
 {\fontsize{10}{12}\selectfont Method} & \textbf{rmCE} & Gauss & Shot & Impulse & Defoc & Glass & Motion & Zoom & Snow & Frost & Fog & Bright & Cont & Elastic & Pixel & JPEG\\
 \midrule
 EMP (BN+O) & $\textbf{97.4}$ & $98.8$ & $98.4$ & $98.8$ & $97.9$ & $99.4$ & $\textbf{96.1}$ & $98.6$ & $101$ & $104$ & $102$ & $97.7$ & $102$ & $\textbf{92.2}$ & $84.8$ & $\textbf{89.5}$\\
 EMP (BN) & $99.0$ & $100$ & $100$ & $99.9$ & $\textbf{97.3}$ & $101$ & $99.2$ & $98.3$ & $102$ & $103$ & $102$ & $109$ & $102$ & $99.6$ & $\textbf{83.1}$ & $89.9$ \\
 ECMP (BN+O) & $102$ & $104$ & $103$ & $104$ & $101$ & $101$ & $98.0$ & $98.3$ & $\textbf{98.5}$ & $101$ & $110$ & $105$ & $106$ & $97.1$ & $97.4$ & $101$\\
 ECMP (BN) & $101$ & $101$ & $99.2$ & $100$ & $98.3$ & $99.5$ & $97.4$ & $\textbf{98.2}$ & $101$ & $\textbf{99.7}$ & $108$ & $113$ & $104$ & $98.7$ & $92.7$ & $97.2$\\
 Dropout & $100$ & $\textbf{98.7}$ & $\textbf{98.1}$ & $\textbf{97.0}$ & $100$ & $\textbf{96.4}$ & $98.9$ & $101$ & $101$ & $99.9$ & $106$ & $114$ & $102$ & $97.1$ & $93.8$ & $94.8$\\
 Vanilla & $100$ & $100$ & $100$ & $100$ & $100$ & $100$ & $100$ & $100$ & $100$ & $100$ & $\textbf{100}$ & $100$ & $100$ & $100$ & $100$ & $100$\\
 \bottomrule
\end{tabular}
\end{table}

\subsection{CIFAR-100}
CIFAR-100 is another commonly used dataset for benchmarking new architectures and algorithms in deep learning \citep{krizhevsky2009learning}. It contains images of size $32$ by $32$ pixels. The PyramidNet we used has the following configuration: depth$=110$, $\alpha=270$, no-bottleneck. It was trained for $400$ epochs with batch size $256$ and tested with $200$ MC samples. We apply our methods based on the PyramidNet/CIFAR-100 implementation provided by the authors \citet{han2017deep} at \href{https://github.com/dyhan0920/PyramidNet-PyTorch}{https://github.com/dyhan0920/PyramidNet-PyTorch}, which uses a standard data augmentation process involving horizontal flipping and random cropping with padding.

Both the BN and output layers were patched in EMP/ECMP for our experiments, given that the parameter overhead for both layers combined is very slight ($1.10\%$). The dropout rate is $0.005$ and a dropout layer was placed before every BN layer.

\subsection{Choosing the MC Sample Size $S$ for Testing}
The number of MC samples $S$ used for testing is also the number of forward passes that the network has to use to evaluate a given data point. We assume that the BN and output layers are patched for EMP/ECMP and evaluate our trained models with $1,20,100,\text{and }200$ samples. Generally, we observe that a increase in $S$ results in better calibration and higher test accuracy. We see in Table \ref{table:mc_size_emp} and Table \ref{table:mc_size_ecmp} that there is no significant difference in ECE between $S=100$ and $S=200$, with $S=200$ performing slightly better in some cases, due to noise in the MC sampling process. This suggests that the additional benefit of using more samples past $S=100$ is slight at best. Another interesting observation is that on PyramidNet, dropout seems to confer little to no advantage in calibration.

\begin{table}[ht]
\vspace{-0.5em}
\caption{Effect of MC Sample Size $S$ on Test Accuracy and Calibration for EMP}
\label{table:mc_size_emp}
\centering
\footnotesize
\begin{tabular}{p{1.1cm} p{1.1cm} p{1.1cm} p{1.1cm} p{1.1cm} p{1.1cm} p{1.1cm} p{1.1cm} p{1.1cm}}
 \toprule
 & \multicolumn{4}{c}{\textbf{ResNet/ImageNet}} & \multicolumn{4}{c}{\textbf{PyramidNet/CIFAR-100}}\\ 
 $S$ & Top-5 & Top-1 & ECE & MCE & Top-5 & Top-1 & ECE & MCE\\
 \midrule
 $1$ & $86.4\%$ & $65.6\%$ & $7.03\%$ & $13.8\%$ & $90.9\%$ & $72.2\%$ & $21.2\%$ & $50.1\%$ \\
 $20$ & $87.1\%$ & $66.9\%$ & $4.08\%$ & $7.00\%$ & $91.6\%$ & $73.2\%$ & $14.2\%$ & $30.0\%$\\
 $100$ & $87.2\%$ & $67.0\%$ & $3.88\%$ & $7.71\%$ & $91.7\%$ & $74.2\%$ & $13.6\%$ & $30.3\%$\\
 $200$ & $87.2\%$ & $67.0\%$ & $3.91\%$ & $6.83\%$ & $92.0\%$ & $74.0\%$ & $13.6\%$ & $28.9\%$\\
 Vanilla & $86.7\%$ & $66.1\%$ & $8.09\%$ & $14.2\%$ & $90.9\%$ & $72.6\%$ & $21.6\%$ & $54.6\%$\\
 \bottomrule
\end{tabular}
\end{table}
\normalsize


\begin{table}[ht]
\vspace{-0.5em}
\caption{Effect of MC Sample Size $S$ on Test Accuracy and Calibration for ECMP}
\label{table:mc_size_ecmp}
\centering
\footnotesize
\begin{tabular}{p{1.1cm} p{1.1cm} p{1.1cm} p{1.1cm} p{1.1cm} p{1.1cm} p{1.1cm} p{1.1cm} p{1.1cm}}
 \toprule
 & \multicolumn{4}{c}{\textbf{ResNet/ImageNet}} & \multicolumn{4}{c}{\textbf{PyramidNet/CIFAR-100}}\\ 
 $S$ & Top-5 & Top-1 & ECE & MCE & Top-5 & Top-1 & ECE & MCE\\
 \midrule
 $1$ & $85.6\%$ & $64.1\%$ & $7.52\%$ & $14.5\%$ & $89.3\%$ & $69.3\%$ & $23.9\%$ & $53.0\%$ \\
 $20$ & $87.1\%$ & $66.9\%$ & $1.70\%$ & $4.18\%$ & $91.6\%$ & $72.9\%$ & $9.35\%$ & $21.4\%$\\
 $100$ & $87.2\%$ & $67.1\%$ & $1.50\%$ & $3.05\%$ & $91.8\%$ & $72.7\%$ & $8.81\%$ & $18.2\%$\\
 $200$ & $87.2\%$ & $67.1\%$ & $1.65\%$ & $3.15\%$ & $91.9\%$ & $73.5\%$ & $8.73\%$ & $18.1\%$\\
 Vanilla & $86.7\%$ & $66.1\%$ & $8.09\%$ & $14.2\%$ & $90.9\%$ & $72.6\%$ & $21.6\%$ & $54.6\%$\\
 \bottomrule
\end{tabular}
\end{table}
\normalsize


\begin{table}[ht]
\vspace{-0.5em}
\caption{Effect of MC Sample Size $S$ on Test Accuracy and Calibration for Dropout}
\label{table:mc_size_dropout}
\centering
\footnotesize
\begin{tabular}{p{1.1cm} p{1.1cm} p{1.1cm} p{1.1cm} p{1.1cm} p{1.1cm} p{1.1cm} p{1.1cm} p{1.1cm}}
 \toprule
 & \multicolumn{4}{c}{\textbf{ResNet/ImageNet}} & \multicolumn{4}{c}{\textbf{PyramidNet/CIFAR-100}}\\ 
 $S$ & Top-5 & Top-1 & ECE & MCE & Top-5 & Top-1 & ECE & MCE\\
 \midrule
 $1$ & $86.3\%$ & $65.3\%$ & $8.99\%$ & $16.6\%$ & $91.0\%$ & $72.0\%$ & $21.8\%$ & $54.2\%$ \\
 $20$ & $86.6\%$ & $65.9\%$ & $7.64\%$ & $14.1\%$ & $91.4\%$ & $72.3\%$ & $21.0\%$ & $50.7\%$\\
 $100$ & $86.7\%$ & $65.9\%$ & $7.64\%$ & $14.2\%$ & $91.1\%$ & $72.2\%$ & $21.0\%$ & $51.9\%$\\
 $200$ & $86.7\%$ & $65.9\%$ & $7.61\%$ & $14.0\%$ & $91.4\%$ & $72.0\%$ & $21.1\%$ & $52.3\%$\\
 Vanilla & $86.7\%$ & $66.1\%$ & $8.09\%$ & $14.2\%$ & $90.9\%$ & $72.6\%$ & $21.6\%$ & $54.6\%$\\
 \bottomrule
\end{tabular}
\end{table}
\normalsize



\subsection{Predictive Performance on Ten Regression Datasets}
\label{section:ten_regression}
This experiment tests the predictive performance of Bayesian neural networks on a collection of ten regression datasets. (The collection of datasets for this purpose was first proposed by \citet{hernandez2015probabilistic}, and later followed by several other authors in the Bayesian deep learning literature \citep{gal2016dropout,louizos2016structured,lakshminarayanan2017simple}.) Each dataset is split $90$:$10$ randomly into training and test sets. Twenty random splits are done (except $Year$ and $Protein$ which uses one and five splits respectively). The average test performance for ECMP, EMP, dropout, and an explicit ensemble is reported in Table \ref{table:ten_regression_1} and Table \ref{table:ten_regression_2}.

Unlike an experiment on a normal dataset with a train-val-test split, \citet{hernandez2015probabilistic}'s experimental setup uses repeated subsampling cross-validation. For each split, the hyperparameters have to be chosen without looking at the test set. While \citet{hernandez2015probabilistic} and \citet{gal2016dropout} use Bayesian optimization to select hyperparameters, it is important to note that the search range of hyperparameters used for different datasets are different, and was determined by looking at the data. (For example, see these \href{https://github.com/yaringal/DropoutUncertaintyExps/blob/a6259f1db8f5d3e2d743f88ecbde425a07b12445/bostonHousing/net/config.pb}{two} \href{https://github.com/yaringal/DropoutUncertaintyExps/blob/a6259f1db8f5d3e2d743f88ecbde425a07b12445/naval-propulsion-plant/net/config.pb}{different} hyperparameter search configurations in \citet{gal2016dropout}'s code repository.) \citet{louizos2016structured} and \citet{lakshminarayanan2017simple} conduct the experiment without using a validation set at all.

We choose to forgo this exercise in hyperparameter tuning, and use fixed hyperparameters throughout. As such, our results are not directly comparable with the results in the literature. It is possible that MC dropout and the explicit ensemble might have significantly different performance under different hyperparameter settings. We do not recommend that others use this experiment as a benchmark, because the experimental setup is fundamentally flawed, as was explained above.

The data in the training set is normalized to have zero mean and unit variance. The neural network used has the ReLU activation function, and one hidden layer with $50$ hidden units, except $Year$ and $Protein$ where we use $100$ hidden units. The BN/dropout layers are placed after the input and after the hidden layer, and only the BN layers ($\gamma$ initialized at $0.2$) are model patched. The networks in the explicit ensemble do not contain any BN or dropout layers. We train the network for $4000$ epochs across all the methods with a batch size of $100$, $\tau_{output} = 0.1$, and weight decay of $0.01$. Where applicable, the dropout rate is $0.005$, ensemble size $K=5$, number of MC samples used $S=10,000$ (same setting as in MC dropout \citep{gal2016dropout}).

\textbf{We observe that ECMP and EMP  have the lowest test root mean squared error in eight of the ten datasets, and the highest test log likelihood in nine of them.}

ECMP and EMP did  worse than dropout and the explicit ensemble  in the $Year$ dataset. We think that this is probably caused by the poor performance of BN on layers that are excessively large compared to the batch size. The \textit{Kin8nm} and \textit{Naval} datasets likely have $\tau_{output} = 0.1$ in the wrong scale, which explains why all the methods show similar results for these two datasets.

\begin{table*}[p]
\caption{\footnotesize Predictive Performance on Ten Regression Datasets (Root Mean Squared Error)}
\label{table:ten_regression_1}
\centering
\begin{tabular}{p{1.3cm} p{1cm} p{1.3cm} p{1.2cm} p{1.2cm} p{1.2cm} p{1.4cm} }
 \toprule
 & & & \multicolumn{4}{c}{Avg.\ Test RMSE and Std.\ Error}\\
 \textbf{Dataset} & Size & Features,\newline Targets & \textbf{ECMP} & \textbf{EMP} & \textbf{Dropout} & \textbf{Ensemble}\\
 \midrule
 Boston & 506 & 13, 1 & \textbf{3.48\newline$\pm$0.18} & 3.56\newline$\pm$0.22 & 3.97\newline$\pm$0.26 & 4.29\newline$\pm$0.27\\
 Concrete & 1,030 & 8, 1 & \textbf{5.61\newline$\pm$0.14} & 5.64\newline$\pm$0.15 & 7.06\newline$\pm$0.19 & 8.81\newline$\pm$0.15\\
 Energy & 768 & 8, 2 & 1.35\newline$\pm$0.06 & \textbf{1.24\newline$\pm$0.04} & 2.63\newline$\pm$0.05 & 3.40\newline$\pm$0.31\\
 Kin8nm & 8,192 & 8, 1 & \textbf{0.08\newline$\pm$0.00} & \textbf{0.08\newline$\pm$0.00} & \textbf{0.08\newline$\pm$0.00} & \textbf{0.08\newline$\pm$0.00}\\
 Naval & 11,934 & 16, 2 & \textbf{0.00\newline$\pm$0.00} & \textbf{0.00\newline$\pm$0.00} & 0.01\newline$\pm$0.00 & 0.01\newline$\pm$0.00\\
 Power & 9,568 & 4, 1 & 4.24\newline$\pm$0.05 & 4.29\newline$\pm$0.05 & 4.07\newline$\pm$0.04 & \textbf{4.04\newline$\pm$0.04}\\
 Protein & 45,730 & 9, 1 & \textbf{1.95\newline$\pm$0.06} & 2.00\newline$\pm$0.07 & 2.01\newline$\pm$0.07 & 2.24\newline$\pm$0.06\\
 WineRed  & 1,599 & 11, 1 & \textbf{0.60\newline$\pm$0.02} & 0.62\newline$\pm$0.02 & 0.63\newline$\pm$0.01 & 0.88\newline$\pm$0.06\\
 Yacht & 308 & 6, 1 & \textbf{1.59\newline$\pm$0.23} & 1.60\newline$\pm$0.28 & 12.90\newline$\pm$1.26 & 29.48\newline$\pm$5.14\\
 Year & 515,345 & 90, 1 & 10.27\newline$\pm$N/A & 12.50\newline$\pm$N/A & \textbf{8.47\newline$\pm$N/A} & 8.69\newline$\pm$N/A\\
 \bottomrule
\end{tabular}
\end{table*}

\begin{table*}[p]
\caption{\footnotesize Predictive Performance on Ten Regression Datasets (Log Predictive Density)}
\label{table:ten_regression_2}
\centering
\begin{tabular}{p{1.3cm} p{1cm} p{1.3cm} p{1.2cm} p{1.2cm} p{1.2cm} p{1.4cm} }
 \toprule
 & & & \multicolumn{4}{c}{Avg.\ Test LPD and Std.\ Error}\\
 \textbf{Dataset} & Size & Features,\newline Targets & \textbf{ECMP} & \textbf{EMP} & \textbf{Dropout} & \textbf{Ensemble} \\
 \midrule
 Boston & 506 & 13, 1 & \textbf{-2.65\newline$\pm$0.05} & -2.70\newline$\pm$0.07 & -2.92\newline$\pm$0.11 & -2.81\newline$\pm$0.07\\
 Concrete & 1,030 & 8, 1 & \textbf{-3.46\newline$\pm$0.07} & -3.59\newline$\pm$0.08 & -4.60\newline$\pm$0.14 & -5.09\newline$\pm$0.13\\
 Energy & 768 & 8, 2  & -2.19\newline$\pm$0.01 & \textbf{-2.15\newline$\pm$0.01} & -2.42\newline$\pm$0.01 & -2.58\newline$\pm$0.03\\
 Kin8nm & 8,192 & 8, 1 & \textbf{-2.07\newline$\pm$0.00} & \textbf{-2.07\newline$\pm$0.00} & \textbf{-2.07\newline$\pm$0.00} & \textbf{-2.07\newline$\pm$0.00}\\
 Naval & 11,934 & 16, 2 & \textbf{-2.07\newline$\pm$0.00} & \textbf{-2.07\newline$\pm$0.00} & \textbf{-2.07\newline$\pm$0.00} & \textbf{-2.07\newline$\pm$0.00}\\
 Power & 9,568 & 4, 1  & -2.95\newline$\pm$0.02 & -2.99\newline$\pm$0.02 & -2.90\newline$\pm$0.01 & \textbf{-2.87\newline$\pm$0.01}\\
 Protein & 45,730 & 9, 1 & \textbf{-2.26\newline$\pm$0.01} & -2.27\newline$\pm$0.01 & -2.27\newline$\pm$0.01 & -2.33\newline$\pm$0.01\\
 WineRed  & 1,599 & 11, 1 & \textbf{-2.09\newline$\pm$0.00} & \textbf{-2.09\newline$\pm$0.00} & \textbf{-2.09\newline$\pm$0.00} & -2.14\newline$\pm$0.01\\
 Yacht & 308 & 6, 1  & -2.25\newline$\pm$0.02 & \textbf{-2.22\newline$\pm$0.05} & -10.79\newline$\pm$1.66 & -5.24\newline$\pm$0.36\\
 Year & 515,345 & 90, 1 & -5.70\newline$\pm$N/A & -6.66\newline$\pm$N/A & -5.66\newline$\pm$N/A & \textbf{-4.47\newline$\pm$N/A}\\
 \bottomrule
\end{tabular}
\end{table*}
\newpage
\section{GPU Memory Analysis Details}
\label{section:gpu_analysis}

In this section, we describe how we created Figure~\ref{fig:gpu_ram}. After examining several websites, we decided to use \citet{gpusite} due to its breadth of information and relatively accurate GPU release dates (specified in days rather than months). We used a series of HTTP requests for different GPU generations to collect all relevant data. After discarding data older than 15 years, we obtained a total of $1599$ unique GPUs. This number is so high because it includes mobile GPUs, desktop GPUs, and workstation GPUs.

We converted the textual representation of each GPU's release date into an integral timestamp, and then plotted this against each GPU's total RAM. A more fine-grained analysis might separate different types of GPUs, or compute the price-per-GB to distinguish inexpensive from high-end GPUs, but we wanted to get an overall idea of the memory trend. It clearly grows exponentially. Fitting the model $[RAM] = 2^{\alpha [Year] + \beta}$ results in $\alpha \approx \frac{1}{2.8}$, meaning the doubling period is every $2.8$ years, but this model visually does not approximate the earlier GPUs very well. We decided to add an additional intercept to fit the model $[RAM] = 2^{\alpha [Year] + \beta} + \gamma$, and here $\alpha \approx \frac{1}{3.2}$, as shown in Figure~\ref{fig:gpu_ram}.

Our analysis is similar to the well-known Moore's Law, which observes that the number of transistors that can be placed in an integrated circuit doubles about every two years. (The transistors also become faster, hence real computing power doubles every 18 months.) Denser silicon can benefit GPU RAM as well, because this memory (DRAM) stores each bit in a single capacitor. Increasing the number of capacitors has a near-linear affect on the amount of bits the RAM can store --- a logarithmic proportion of the silicon must be dedicated to addressing the bits, which are arranged in banks and must be refreshed periodically to prevent capacitors from losing their charge. It is interesting that we observe RAM capacity doubling every three years, somewhat slower than CPU speed increases, but less research effort is likely dedicated to shrinking capacitors compared with transistors. Furthermore, memory requires several layers of cache to be useful (even in GPUs), which requires some processor/motherboard co-design and may also contribute to the longer doubling time.

Although Moore's Law has started to break down recently because physical limits are being reached, the observation that technological capabilities grow exponentially is still sound; research is pushing to use additional silicon for other purposes, such as massively parallel CPU cores and special-purpose hardware (of which GPUs are an early example). As deep learning grows in significance, we are even starting to see special-purpose hardware for it, such as Google's Tensor Processing Units (TPUs). We believe that the pressures of increasingly large models will drive new hardware to include more and more memory. Even if access latency is increased, neural-network hardware may move towards an even deeper memory hierarchy, much as traditional operating systems embrace swap memory to increase their capabilities. In the past few decades, the clock speed of individual cores was the most significant metric of computing progress, but as deep learning and other frontiers of computer science utilize increasing parallelization, we hypothesize that memory capacity will be the more relevant metric in the future of computing.
\end{document}